%% file: main.tex
\PassOptionsToPackage{table,xcdraw,dvipsnames}{xcolor}
\documentclass[]{style}

\usepackage[toc,page,header]{appendix}

\input{resources/packages}
\usepackage{array}
\usepackage{amsthm}

\input{math_commands.tex}

\newcommand{\sectionreducemargin}[1]{\section{#1}}
\newcommand{\subsectionreducemargin}[1]{\subsection{#1}}

\title{Rethinking World-Action Model for Compositional \\ and In-Context Robotic Manipulation}

\author{
  Shukai Gong$^{1*}$\hspace{0.1cm}
  Xuanran Zhai$^{2*}$\hspace{0.1cm}
  Yintianrun Zhang$^{1*}$\hspace{0.1cm}
  Ruopeng Cui$^{2}$\hspace{0.1cm}
  Ye Huang$^{1}$\hspace{0.1cm}
  Yiyang Fu$^{1}$\hspace{0.1cm}
  Dexuan Lyu$^{2}$\hspace{0.1cm}
  Chaojie Li$^{2}$\hspace{0.1cm}
  Xinyi Song$^{2}$\hspace{0.1cm}
  Peiwen Lin$^{2}$\hspace{0.1cm}
  Chuang Wang$^{2}$\hspace{0.1cm}
  Mingyuan Jia$^{3}$\hspace{0.1cm}
  Yufan Deng$^{1}$\hspace{0.1cm}
  Jiaxin Fang$^{3}$\hspace{0.1cm}
  Bo Liang$^{1}$\hspace{0.1cm}
  Jiaxin Li$^{1}$\hspace{0.1cm}
  Yuxiang Gao$^{3\dagger}$\hspace{0.1cm}
  Hao Liu$^{2\dagger}$\hspace{0.1cm}
  Daquan Zhou$^{1\dagger}$\hspace{0.1cm}
}

\affiliation[1]{Peking University}
\affiliation[2]{AgiBot}
\affiliation[3]{CocoMatrix}
\contribution[*]{Equal contribution}
\contribution[\dagger]{Corresponding author}

\input{Secs/0_abs}

\checkdata[
\raisebox{-0.2em}{\includegraphics[width=0.025\linewidth]{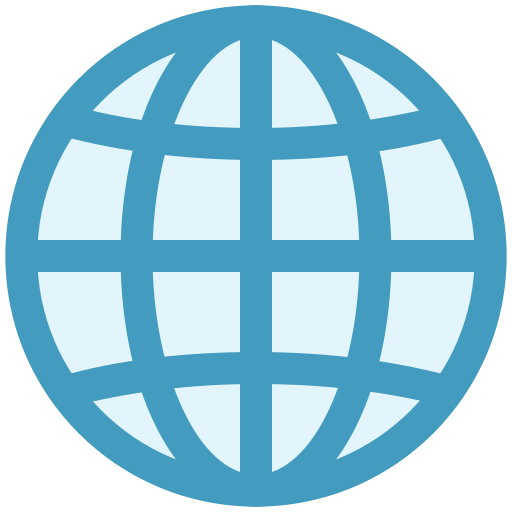}}~~Project Page]{\href{https://dagroup-pku.github.io/ViGAR}{\texttt{https://dagroup-pku.github.io/ViGAR}}
\\[-1.5ex]}

\checkdata[
\raisebox{-0.2em}{\includegraphics[width=0.025\linewidth]{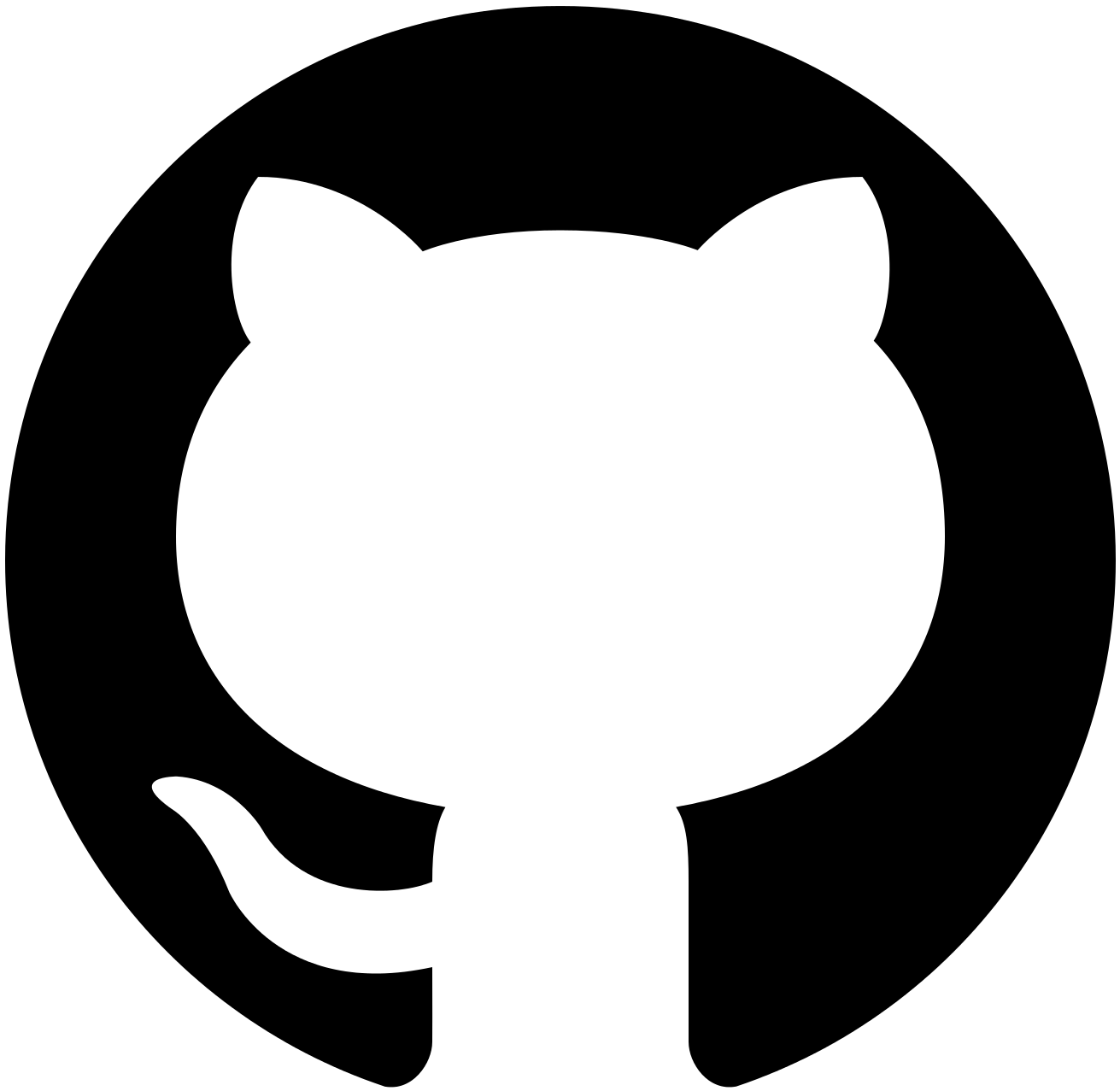}}~~GitHub Repo]{\href{https://github.com/DAGroup-PKU/ViGAR}{\texttt{https://github.com/DAGroup-PKU/ViGAR}}
\\[-1.7ex]}

\checkdata[
\raisebox{-0.2em}{\includegraphics[width=0.025\linewidth]{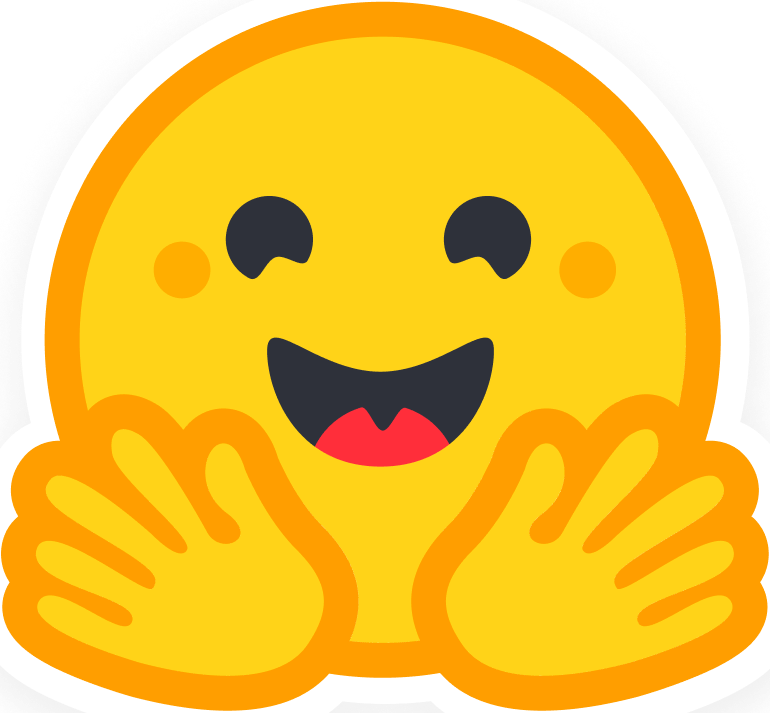}}~~Huggingface Repo]{\href{https://huggingface.co/DAGroup-PKU/ViGAR}{\texttt{https://huggingface.co/DAGroup-PKU/ViGAR}}
\\[-1.7ex]}

\crefname{figure}{Fig.}{Figs.}
\crefname{table}{Table}{Tables}
\crefname{section}{Section}{Sections}
\crefname{equation}{Equation}{Equations}

\begin{document}

\maketitle

\begin{figure}[h]
    \centering
    \includegraphics[width=0.97\columnwidth]{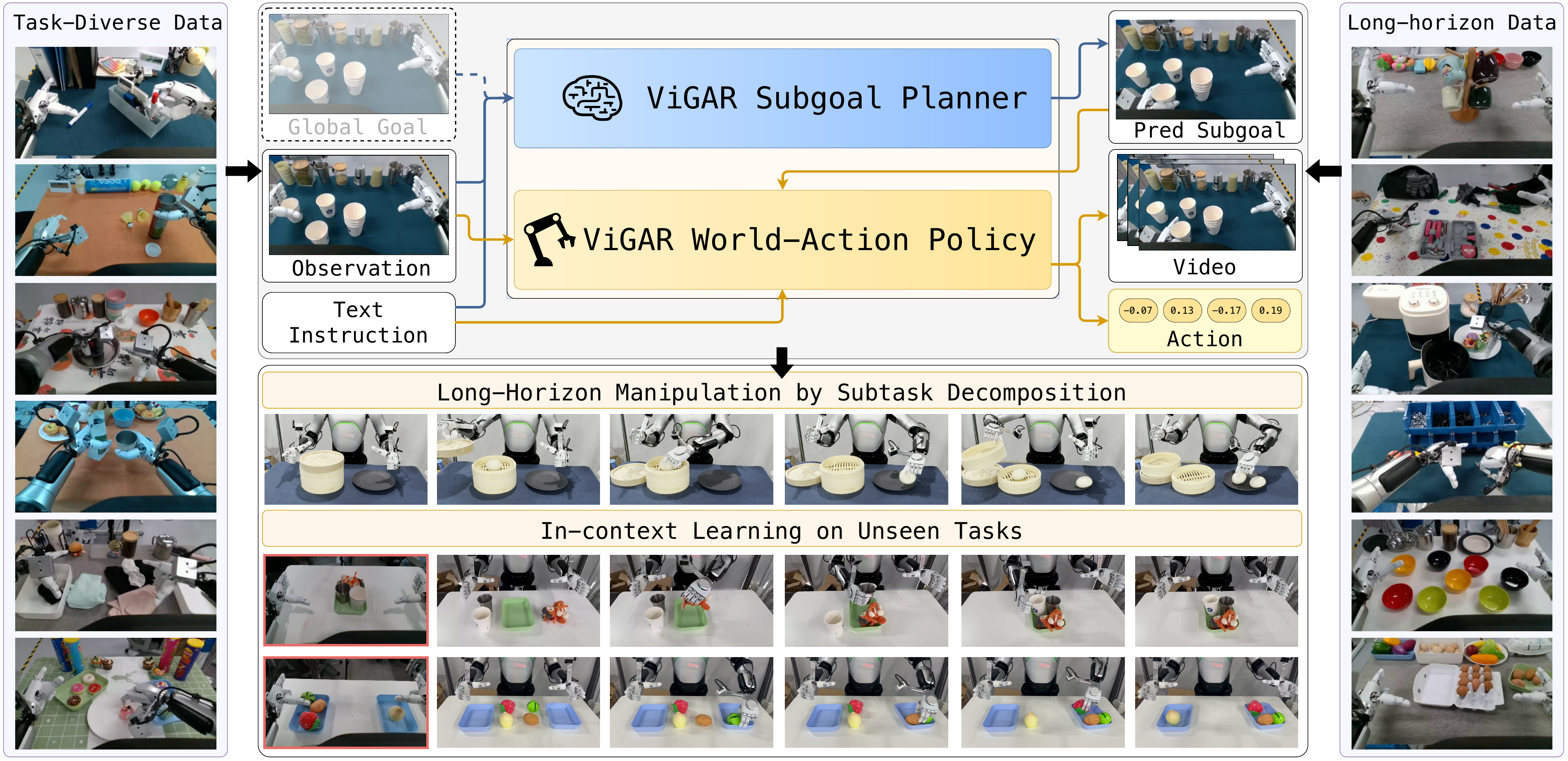}
    \caption{
    \textbf{Overview of ViGAR.} Our framework factorizes long-horizon compositional manipulation tasks into visual subgoal planning and subgoal execution, with both modules trained on task-diverse, long-horizon robot data. ViGAR achieves strong compositional manipulation performance in both simulated and real-world settings, and generalizes to unseen tasks through in-context steering.
    }
    \label{Fig: teaser}
\end{figure}

\input{Secs/1_intro}
\input{Secs/2_related}
\input{Secs/3_prelim}
\input{Secs/4_method}
\input{Secs/6_experiments}
\input{Secs/7_conclusion}

\bibliographystyle{plainnat}
\setlength{\bibhang}{0pt}
\setlength\bibindent{0pt}
\newpage
\bibliography{iclr2027_conference}

\newpage
% \appendix

\input{Secs/8_appendix}

\end{document}

%% file: resources/packages.tex
\usepackage{natbib}

\usepackage{CJKutf8}

\usepackage{xargs}  

\usepackage{todonotes}  

\usepackage{multirow}

\usepackage{cleveref}

\usepackage{amsmath}
\usepackage{dsfont}

\usepackage{svg}

\usepackage{mathrsfs}
\usepackage{adjustbox}
\usepackage{multirow}
\usepackage{multicol}
\usepackage{tcolorbox}
\usepackage{changepage}
\usepackage{enumitem}
\usepackage{graphicx}
\usepackage{amssymb}
\usepackage{xcolor}
\usepackage{float}
\usepackage{multirow}
\usepackage{threeparttable}
\usepackage{graphicx}
\usepackage{subcaption}
\usepackage{algorithm}
\usepackage{algpseudocode}
\usepackage{wrapfig}
\usepackage{epsfig}
\usepackage{overpic}
\usepackage{diagbox}
\usepackage{fontawesome}
\usepackage{bbding}
\usepackage{wasysym}
\usepackage{utfsym}
\usepackage{mathtools}
\usepackage[table]{xcolor}  % loads xcolor with table support
\usepackage{colortbl}       % gives \columncolor, \rowcolor
\usepackage{tabularx} % Add to your preamble
\usepackage{makecell} % Add to your preamble
\usepackage[dvipsnames]{xcolor}

\newcolumntype{g}{>{\columncolor{gray!10}}c} % gray background

\definecolor{catgray}{gray}{0.9}
\definecolor{skyblue}{rgb}{0.53,0.81,0.92} % sky blue

\colorlet{skyblue!30}{skyblue!30!white} % 30% skyblue, 70% white

\definecolor{customblue}{RGB}{70,130,180}  % This is equivalent to rgb(70,130,180)

\newtcolorbox{evolbox}[2][]{%
  enhanced,
  colframe=customblue,
  colback=white,
  coltitle=white,
  rounded corners,
  boxrule=1pt,
  titlerule=0pt,
  toptitle=1mm,
  bottomtitle=1mm,
  fonttitle=\bfseries,
  width=#2\textwidth, % This takes the second parameter as the width fraction
  #1
}
\usepackage{minitoc}
\usepackage{url}

\PassOptionsToPackage{table,xcdraw}{xcolor}
\usepackage{titletoc}
\usepackage{placeins}
\usepackage{pifont}

\definecolor{RowBlue}{HTML}{E9F2FB}
\definecolor{RowRed}{HTML}{F9EAEA}
\definecolor{Top1}{HTML}{50DB4B} % 深绿
\definecolor{Top2}{HTML}{A5FFA2} % 中绿
\definecolor{Top3}{HTML}{D9FFD9} % 浅绿
\definecolor{Sub1}{HTML}{EAB8B8}
\definecolor{Sub2}{HTML}{E4E4E4}

\renewcommand{\emph}[1]{\textit{#1}}

%% file: math_commands.tex
\usepackage{amsmath,amsfonts,bm}

\def\eqref#1{equation~\ref{#1}}
\def\1{\bm{1}}

\DeclareMathAlphabet{\mathsfit}{\encodingdefault}{\sfdefault}{m}{sl}
\SetMathAlphabet{\mathsfit}{bold}{\encodingdefault}{\sfdefault}{bx}{n}

%% file: Secs/0_abs.tex
\abstract{
Long-horizon compositional manipulation has become increasingly important for real-world robot deployment, where a single task involves multiple coordinated subtasks. Existing world-action models (WAMs) jointly predict short-horizon visual futures and actions, but typically lack explicit subtask-level reasoning. We propose \textbf{Visual Goal-conditioned Action Reasoning (ViGAR)}, a hierarchical framework that factorizes manipulation into 
a visual subgoal planner and a subgoal executor.
Given the current observation and global instruction, the subgoal planner predicts a visual subgoal for the next subtask. The subgoal executor then jointly generates future visual trajectories and actions conditioned on the predicted subgoal. Both components share a pretrained world-model representation, enabling task-level planning and action generation to benefit from common physical knowledge. Moreover, our framework naturally supports in-context learning: using a global goal image as context can induce different subtask decompositions and behaviors without parameter updates. 
On the RoboTwin Clean2Random benchmark, ViGAR achieves \textbf{82.00\%} and \textbf{67.02\%} success rates under the \emph{Clean} and \emph{Random} settings, respectively, surpassing the strongest baseline by 12.86 percentage points in average success rate.
Real-world robot experiments on five compositional and two in-context learning tasks further confirm the effectiveness of ViGAR.
}

%% file: Secs/1_intro.tex
\sectionreducemargin{Introduction} \label{Sec: intro}

Robots in production and service settings increasingly face compositional manipulation tasks that involve multiple coordinated subtasks. Such tasks require policies to maintain progress across subtask transitions while generalizing to unseen scenes, objects, and spatial layouts, rather than replaying a fixed sequence from training.
Policies for real-world deployment therefore need both embodied generalization and mechanisms for organizing behavior over extended task horizons.
World-Action Models (WAMs) offer an appealing paradigm for this goal. By jointly predicting future visual states and actions, WAMs learn physically grounded action dynamics from dense video supervision and have demonstrated strong generalization across tasks, environments, embodiments, and visual distribution shifts~\citep{chen2025large, li2026causal, ye2026world, agarwal2026cosmos}.

Despite these strengths, existing WAMs are not explicitly designed for long-horizon compositional manipulation. WAM formulations typically provide a useful local control signal by modeling what is likely to happen next, but do not explicitly organize behavior across a sequence of subtasks or expose a task-level interface through which the desired progression can be specified.
Recent studies~\citep{black2024zero,zhu2026dswam,intelligence2025pi_} reveal that long-horizon task progression requires more than supervision on short-term evolution of the physical world, and further demonstrate the value of predicting semantic subtasks for open-world long-horizon manipulation and generalization. The core challenge is therefore to endow WAMs with explicit task-level decision making while preserving their strong world-centric prediction and physical generalization.

One line of work externalizes this decision making to a language planner that exploits VLMs to decompose a task into linguistic subtasks for a policy to execute.
However, language subtask descriptions often underspecify precise spatial outcome without ambiguity, whereas visual subgoals impose stronger constraints on the desired scene configuration~\citep{intelligence2026pi}. Our key insight is that compositional manipulation can be factorized into \emph{long-term future selection} and \emph{short-term future realization}. World-action modeling for long-horizon tasks should first make explicit the target state robot intends to realize, and then generate the actions that will bring it about. 

To this end, we propose \textbf{Visual Goal-conditioned Action Reasoning (ViGAR)}, which treats subgoal images as explicit decision variable linking task-level reasoning to physical action generation. 
Given the current observation and the global task instruction, ViGAR first performs visual goal reasoning to predict the desired subtask goal image. 
Then, conditioned on the current observation, global task instruction and predicted subgoal image, it jointly predicts future visual trajectories and corresponding robot actions leading to the subgoal.
Both task-level decision and physical realization are learned in one shared generative world-model representation, so that the model can leverage the same physical knowledge for both stages.

Moreover, our framework naturally supports in-context learning: with the scene and instruction held fixed, changing only the global goal induces a different composition of behaviors, forcing the policy to reorganize its action sequence rather than default to the most likely training trajectory. We term this \emph{in-context robotic manipulation}: the global goal image serves as context for inferring a subtask decomposition, so an out-of-distribution global goal elicits a task composition unseen in training, with no parameter update.

Experiments on both simulation benchmark and real-robot compositional tasks validate the effectiveness of ViGAR. On the RoboTwin Clean2Random benchmark~\citep{chen2025robotwin}, ViGAR achieves success rates of \textbf{82.00\%} and \textbf{67.02\%} under the \emph{Clean} and \emph{Random} settings, respectively, surpassing the strongest baseline~\citep{yang20264d} by 12.86 percentage points in average success rate. 
In real-robot experiments, ViGAR outperforms strong VLA and Cosmos3-Nano-Policy baselines on five compositional tasks, and exhibits in-context learning ability on unseen global goals, confirming the validity of our method.
Our contributions are summarized as follows:
% \vspace{-4pt}
\begin{itemize}
    \item We factorize compositional manipulation into \textbf{long-term future selection} and \textbf{short-term future realization}, and propose \textbf{ViGAR}, a hierarchical framework that predicts the next subgoal as an explicit decision variable and conditions joint video-action generation on it.
    \item We identify \textbf{in-context robotic manipulation} as a capability that emerges from explicit goal conditioning. With the scene and instruction held fixed, a new global goal image elicits a different subtask decomposition and a correspondingly reorganized action sequence, specifying out-of-distribution task compositions at deployment without any parameter update.
    \item ViGAR outperforms strong baselines on both simulation benchmark and real-robot compositional manipulation tasks, and exhibits strong in-context learning ability on unseen global goals, confirming the validity of our approach.
\end{itemize}

%% file: Secs/2_related.tex
\sectionreducemargin{Related Works} \label{Sec: related works}

\textbf{VLAs for Compositional Robotic Manipulation.} Vision-language-action (VLA) models map visual observations and language instructions to continuous action trajectories, and have demonstrated strong generalizability across objects, scenes, and tasks~\citep{brohan2023rt, black2024pi_0, intelligence2025pi_,rdt2,ZhaiX-RSS-26,internvla,fu2026stablevla,qu2025spatialvla}. 
For compositional manipulation, however, a direct instruction-to-action mapping is often insufficient due to increased complexity, and recent systems introduce subtask information as an explicit planning interface, where a VLM or a pretrained world model synthesizes subtask information that conditions a VLA policy~\citep{intelligence2025pi_, cai2026tau_0, intelligence2026pi, long2026scaling}.

\textbf{World Models for Robotic Manipulation.} World models learn the structure and evolution of physical environments by predicting future states, and world-action models (WAMs) extend this paradigm by jointly modeling future observations and actions.
Recent studies have demonstrated that WAMs exhibit strong generalizability beyond the training domain, owing to their explicit modeling of geometric and temporal transitions~\citep{li2026causal, ye2026world, agarwal2026cosmos,ma2026humanscaleegocentrichumanvideo}.
However, WAMs typically focus on local action generation and lack a mechanism for reasoning about the plan a compositional task requires.
BagelVLA~\citep{hu2026bagelvla} and RxBrain~\citep{liang2026rxbrain} generate interleaved subtask text and subgoal images within a unified model, but these predictions either guide action generation implicitly through intermediate features or serve primarily as planning outputs, rather than acting as explicit subgoal conditions for the downstream policy.

\textbf{In-Context Learning for Robotic Manipulation.}
In large language models, a novel task can be performed simply by specifying it in the context without any parameter update, termed \emph{in-context learning (ICL)}~\citep{brown2020language, wei2022emergent, wei2022chain}.
Carried into robotic manipulation, this moves the acquisition of a new task out of training-time imitation and into inference time, bypassing the need for costly data collection and model fine-tuning.
Zero-WAM~\citep{zhou2026zero} performs ICL by prompting a causal video-action model with human demonstration videos; HOST~\citep{chen2026robots} acquires a new skill from a single human video by predicting the robot's own future observations along the demonstrated progression; Skild S1~\citep{skild2026s1} and GEN-1.5~\citep{generalist2026gen15} likewise condition on a demonstration video or a short sensorimotor trajectory. In each case the context conveys \emph{how} a task is performed, and the policy re-executes it under its own embodiment.
We instead specify a task by its terminal world state: a global goal image constrains only \emph{what} the scene should become, leaving the subtask decomposition to the model.

%% file: Secs/3_prelim.tex
\sectionreducemargin{Preliminaries} \label{Sec: prelim}
\subsectionreducemargin{VLA and WAM for Robotic Manipulation}
Consider a robot task specified by a language instruction $\ell$. At inference step $t$, a VLA policy $\pi^{\text{VLA}}_\theta$ with parameters $\theta$ predicts an $H_a$-step action chunk $\mathbf{A}_{t+1}:=\mathbf{a}_{t+1:t+H_a}\in \mathbb{R}^{H_a\times d_a}$ conditioned on the current multi-view observation $\mathbf{o}_t = \{\mathbf{o}_t^v\}_{v=1}^{V}$, proprioceptive state $\mathbf{s}_t \in \mathbb{R}^{d_s}$, and textual metadata $\mu$ specifying the embodiment, camera resolution, and control mode:
\begin{equation}
    \mathbf{A}_{t+1} \sim \pi^{\text{VLA}}_\theta(\mathbf{A}_{t+1}|\mathbf{o}_t,\mathbf{s}_t,\ell,\mu),\notag
\end{equation}
where $H_a$ denotes the action chunk size. A WAM policy $\pi^{\text{WAM}}_\psi$ with parameters $\psi$ is similarly conditioned on $\mathbf{o}_t, \mathbf{s}_t, \ell, \mu$, but instead parameterizes a joint conditional distribution over a short visual future $\mathbf{O}_{t+1}:=\mathbf{o}_{t+1:t+H_v}$ and an action chunk $\mathbf{A}_{t+1}$:
\begin{equation}
    (\mathbf{O}_{t+1},\mathbf{A}_{t+1}) \sim \pi_\psi^{\text{WAM}} (\mathbf{O}_{t+1},\mathbf{A}_{t+1}|\mathbf{o}_t,\mathbf{s}_t,\ell,\mu).\notag
\end{equation}

\vspace{-10pt}
\subsectionreducemargin{Problem Formulation for Compositional Robot Tasks}
Let $T\in \mathcal{T}$ denote a compositional robot task specified by a global instruction $\ell\in \mathcal{L}$, where $\mathcal{T}$ and $\mathcal{L}$ denote the space of language-conditioned robot tasks and the space of free-form text instructions, respectively. To address the challenge of compositional manipulation, instead of modeling a policy $\pi(\cdot|\mathbf{o}_t,\mathbf{s}_t,\ell,\mu)$ that generates action based on the global instruction $\ell$, we decompose $T$ into $n$ intermediate subtasks $T_i\in \mathcal{T}$, such that $T=(T_1,\cdots,T_n)$. 
For each subtask $T_i$ at stage $i$, we specify a subtask instruction $\ell_i$ describing the subtask goal in text, and a subgoal image $\mathbf{g}_i\in \mathcal{I}$ specifying the visual target state the robot should reach, where $\mathcal{I}$ is the space of images. The objective is then to train a robot policy $\pi$ that leverages the subtask information $(\ell_i,\mathbf{g}_i)$ to predict the action trajectory for subtask $T_i$:
\begin{equation}
    \mathbf{A}^{T_i}_{t+1}\sim \pi(\cdot|\mathbf{o}_t,\mathbf{s}_t,\ell_i, \mathbf{g}_i,\mu),\ i=1,\cdots,n,\notag
\end{equation}
where $t$ denotes the current timestep. The robot executes $\mathbf{A}^{T_i}_{t+1}$ until the current subtask is completed, at which point the subtask instruction and subgoal image are advanced to $\ell_{i+1}$ and $\mathbf{g}_{i+1}$, respectively. In the following part, we use `subgoal' to term $(\ell_i,\mathbf{g}_i),\ i=1,\cdots,n$.

\subsectionreducemargin{In-Context Task Specification for Robotic Manipulation}
In-context learning (ICL) performs a task specified at deployment rather than acquired during training. In addition to language instruction $\ell$, let $\mathcal{C}$ denote an optional context supplied at inference time. A policy $\pi$ with parameters $\phi$ performs in-context manipulation if a new task is realized by changing $\ell$ or $\mathcal{C}$ alone without the update of $\phi$:
\begin{equation}
    \mathbf{A}_{t+1} \sim \pi_{\phi}(\cdot|\mathbf{o}_t,\mathbf{s}_t,\ell,\mathcal{C},\mu). \label{icl-prelim}
\end{equation}
Recent work~\citep{zhou2026zero, chen2026robots, skild2026s1, generalist2026gen15} on robotic in-context learning often instantiates $\mathcal{C}$ as a demonstration video, such as a human operation video $\mathbf{V}:=\mathbf{v}_{1:K}$, which prescribes the procedure the robot should reproduce.

%% file: Secs/4_method.tex
\sectionreducemargin{Method} \label{Sec: method}
In this section, we describe the two essential parts of ViGAR: a subgoal planner and a subgoal-guided world-action model.
We divide each compositional manipulation task into multiple subtasks, where each subtask contains an atom action and a subgoal $(\ell_i,\mathbf{g}_i)$.
Our subgoal planner generates a subgoal for each subtask, as illustrated in \cref{method_A}. 
Then, our world-action policy predicts the action chunk from the current observation, proprioceptive state and generated subgoal, as detailed in \cref{method_B}. 
Finally, we show in \cref{method_C} that by specifying a global goal as an inference-time context, the subgoal planner can generate a subtask decomposition consistent with the global goal, allowing a new task composition to be specified at deployment without any parameter
updates.

Both the subgoal planner and the downstream policy are initialized from the same pretrained world model~\citep{agarwal2026cosmos}. This backbone follows a dual-branch design, with a \emph{reasoner} branch processing language and semantic visual tokens, and a \emph{generator} branch generating image or video via flow matching. The two branches interact through a shared multimodal attention layer.

\begin{figure}[t]
    \centering
    \includegraphics[width=0.9\columnwidth]{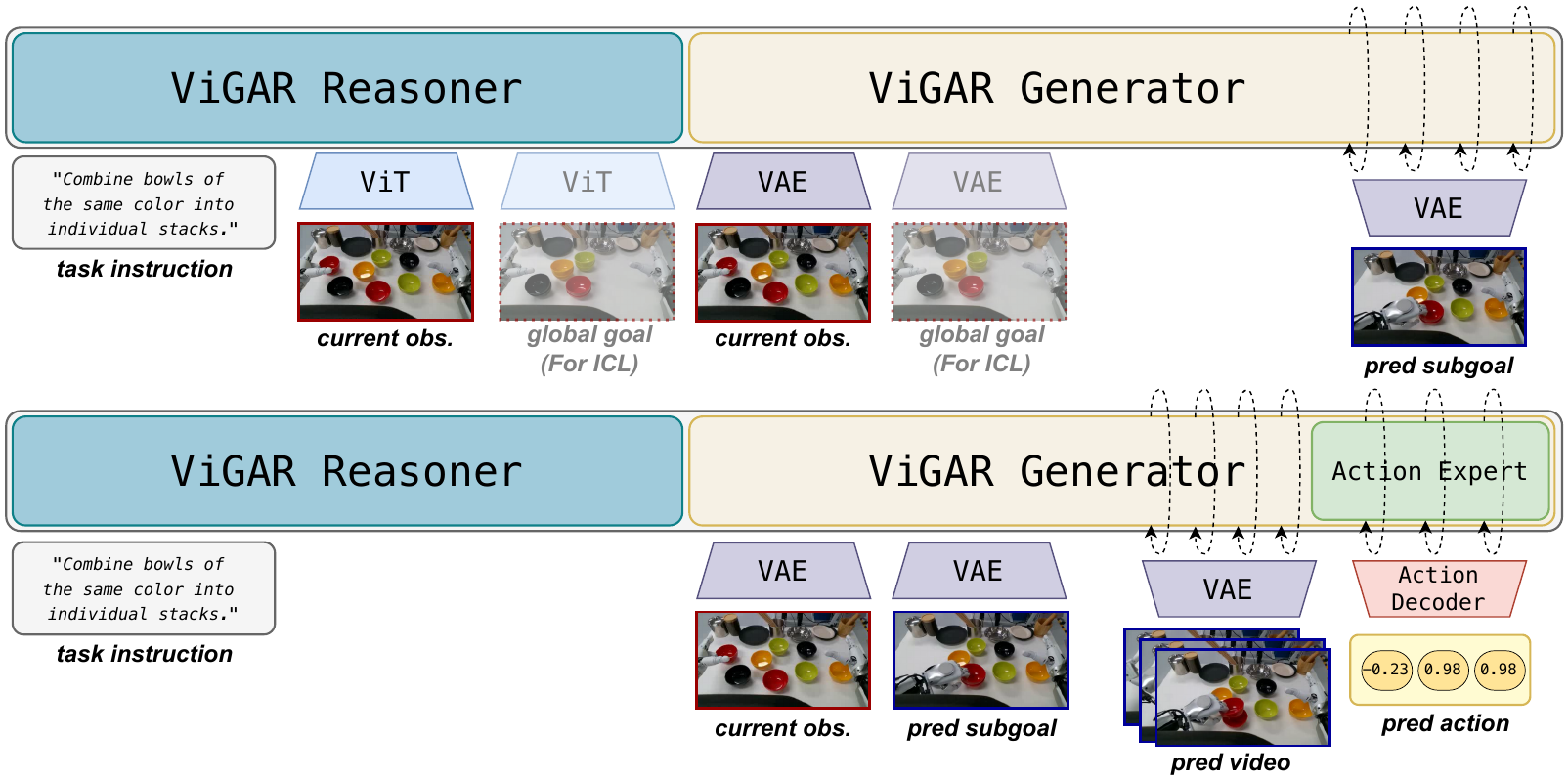}
    \caption{
    \textbf{Model architecture of ViGAR:} (1)~the subgoal planner $\mathcal{G}_\theta$ generates the next subgoal image from the task instruction and current observation, optionally conditioned on a global goal image for in-context learning (ICL); (2)~the policy $\pi^{\text{WAM}}_\phi$ takes in the predicted subgoal and jointly predicts the future frames and the action chunk.  
    }
    \label{fig:arch}
\end{figure}

\subsectionreducemargin{Subgoal Planner for Task-level Reasoning}
\label{method_A}

\begin{figure*}[t]
    \centering
    \begin{subfigure}[b]{0.42\textwidth}
        \centering
        \includegraphics[width=\linewidth]{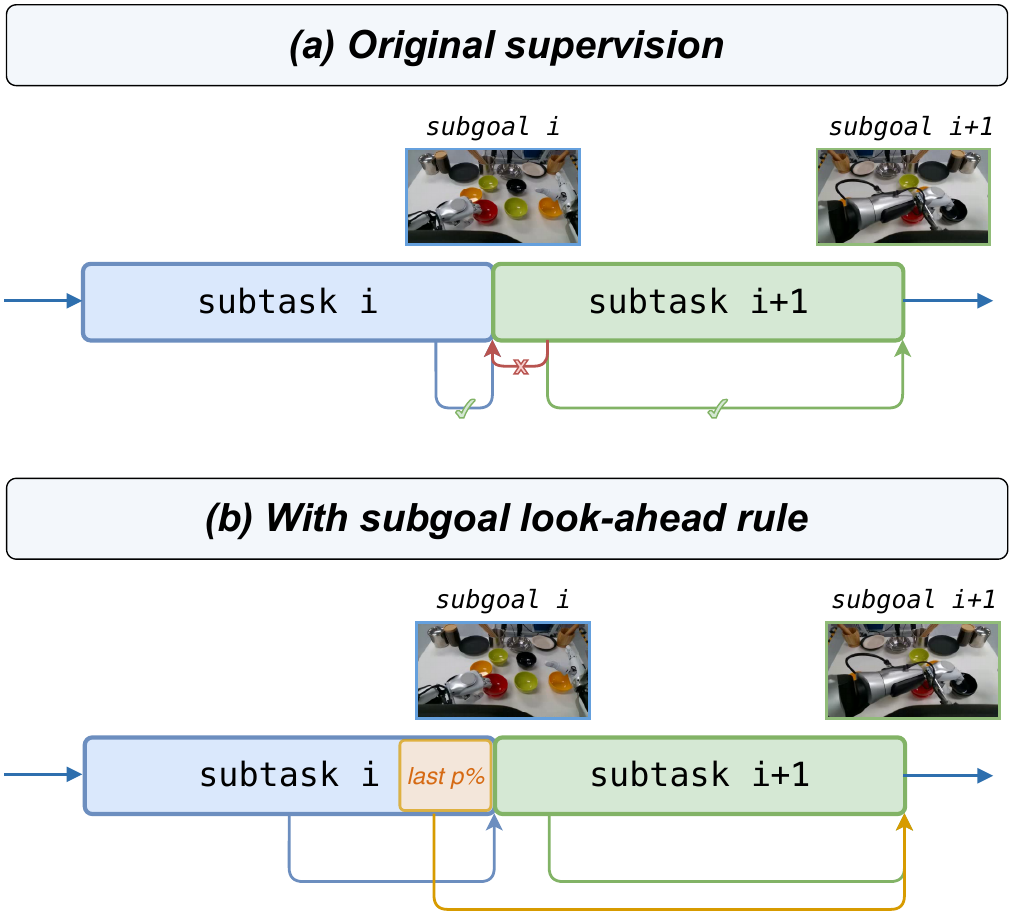}
        \caption{\textbf{Subgoal look-ahead rule} (\cref{look-ahead rule}): redirecting the last $p\%$ of $T_i$ to $\mathbf{g}_{i+1}$ makes the supervision consistent across the boundary.}
        \label{fig:subgoal_skipping}
    \end{subfigure}
    \hfill
    \begin{subfigure}[b]{0.56\textwidth}
        \centering
        \includegraphics[width=\linewidth]{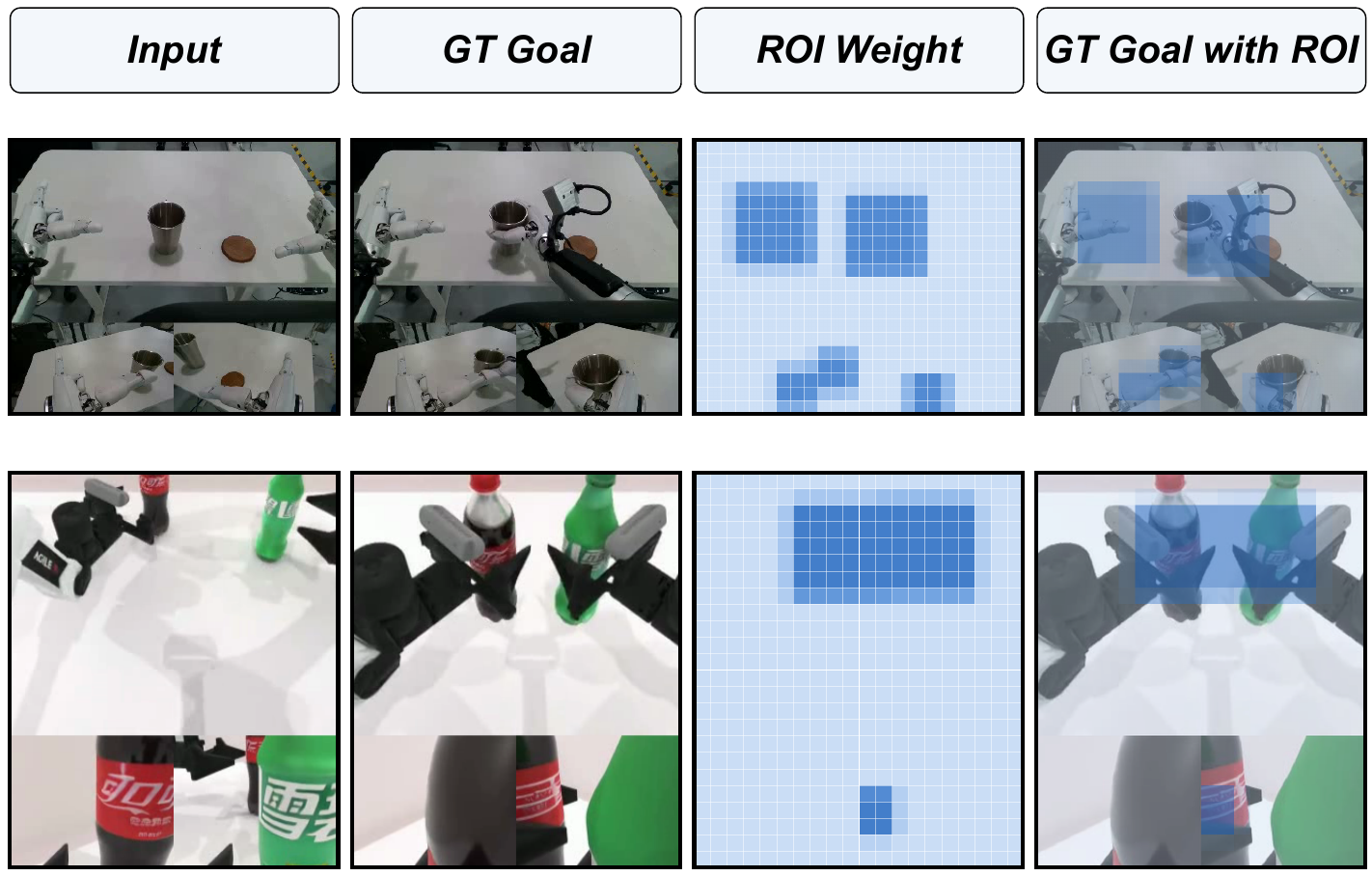}
        \caption{\textbf{End-effector ROI weighting} (\cref{roi-weight}): the planning loss is weighted toward the end-effector region-of-interest (ROI), so that supervision concentrates where manipulation actually happens rather than on static background.}
        \label{fig:subgoal_roi}
    \end{subfigure}
    \caption{Implementation design in subgoal supervision: (a) temporal boundary handling via look-ahead redirection, and (b) spatial weighting via end-effector region-of-interest masks.}
    \label{fig:subgoal_design}
    \vspace{-6pt}
\end{figure*}

In subgoal planning, we fix $\ell_i = \ell$ for all $i$, rather than decomposing the global instruction into stage-specific subtask text, as the subgoal image $\mathbf{g}_i$ already provides precise grounding of the target state, and retaining $\ell$ serves as a consistent semantic anchor for the overall task. This also alleviates potential error accumulation that subtask text generation could introduce to the downstream policy.

Guided by this design, we cast subgoal planning as an image-editing problem:
given a reference frame $\mathbf{o}_t$ within subtask $T_i$ and the global
instruction $\ell$, the subgoal planner $\mathcal{G}_\theta$ predicts a
subgoal image $\mathbf{g}_i$ useful for executing $T_i$ and continuing the
global task $T$. 

Notably, as shown in~\cref{fig:subgoal_skipping}, frame $\mathbf{o}_t$ at the end of $T_i$ can be visually similar to early frames of $T_{i+1}$, while the two are assigned different targets, $\mathbf{g}_i$ and $\mathbf{g}_{i+1}$. 
Without correction, the planner tends to confuse them at inference time, regenerating $\mathbf{g}_i$ once the robot has entered $T_{i+1}$ and driving the plan back to a subtask already completed. To keep supervision consistent for visually similar inputs, we redirect such $\mathbf{o}_t$ to target $\mathbf{g}_{i+1}$ by the following \emph{subgoal look-ahead rule}:
\begin{equation}
    \mathrm{target}(\mathbf{o}_t) =
    \begin{cases}
        \mathbf{g}_{i+1}, & \mathbf{o}_t \in \text{last } p\% \text{ of } T_i,\ i<n \\
        \mathbf{g}_i, & \text{otherwise}.
    \end{cases} \label{look-ahead rule}
\end{equation}
For notation simplicity, we still denote $\mathbf{g}_i$ as the target assigned to $\mathbf{o}_t$ after this subgoal look-ahead rule. Training pairs $(\mathbf{o}_t,\ell,\mathbf{g}_i)$ are constructed from subtask-segmented robot demonstrations, with each subtask corresponding to a demonstration segment ending at its target frame.

Formally, we train $\mathcal{G}_\theta$ with a flow-matching objective. Let $\mathbf{x}_0^g = \mathcal{E}_v(\mathbf{g}_i)$ denote the target latent and $\mathbf{z}_t = \mathcal{E}_v(\mathbf{o}_t)$ denote the reference latent, where $\mathcal{E}_v$ is the VAE encoder. We construct a noisy latent $\mathbf{x}_\tau^g=(1-\tau)\mathbf{x}_0^g+\tau\boldsymbol{\epsilon}$, where $\tau\sim \mathcal{U}[0,1], \boldsymbol{\epsilon}\sim \mathcal{N}(0,\mathbf{I})$, and define the flow-matching target $\mathbf{v}_{\mathrm{plan}} = \boldsymbol{\epsilon}-\mathbf{x}_0^g$.

Because manipulation-related changes are concentrated near the end effector, we construct an end-effector-centered mask to emphasize these regions during training. As shown in~\cref{fig:subgoal_roi}, we project the corresponding expert end-effector positions through the calibrated cameras and take the union of fixed-size windows around visible projections to form a binary mask $\mathbf{m}_i$, and further area-pool it to $\tilde{\mathbf{m}}_i$ under latent resolution. For each target latent token $u$, we define its weight as
\begin{equation}
    w_i(u)=
    \frac{1+\lambda\tilde{m}_i(u)}
    {\frac{1}{|\Omega|}
    \sum_{u'\in\Omega}
    \left(1+\lambda\tilde{m}_i(u')\right)},
    \qquad
    \frac{1}{|\Omega|}\sum_{u\in\Omega}w_i(u)=1.
    \label{roi-weight}
\end{equation}
The subgoal planner is then optimized with the following weighted flow-matching loss:
\begin{equation}
    \mathcal{L}_{\mathrm{Plan}}
    =
    \mathbb{E}_{(\mathbf{o}_t,\ell,\mathbf{g}_i), \tau, \boldsymbol{\epsilon}}
    \left[
    \left\|
    \sqrt{\mathbf{w}_i}\odot
    \left(
    \mathcal{G}_\theta
    \left(\mathbf{x}_\tau^g,\tau\mid\mathbf{z}_t,\ell\right)
    -\mathbf{v}_{\mathrm{plan}}
    \right)
    \right\|_2^2
    \right],
    \label{plan-loss}\notag
\end{equation}
where $\mathbf{w}_i$ collects the weights $w_i(u)$. The predicted subgoal $(\ell,\hat{\mathbf{g}}_i)$ serves as a task-level reasoning proxy for the downstream action policy to complete $T_i$.

\subsectionreducemargin{Subgoal-guided World-Action Model} \label{method_B}

The policy $\pi^{\text{WAM}}_\phi$ predicts a short visual future $\mathbf{O}_{t+1}$ and an action chunk $\mathbf{A}_{t+1}$ conditioned on the current observation $\mathbf{o}_t$, proprioceptive state $\mathbf{s}_t$, and the predicted subgoal $(\ell,\hat{\mathbf{g}}_i)$ of the active subtask $T_i$. It is trained with a joint flow-matching objective over the visual future and the action chunk. Let $\mathbf{x}_0^v = \mathcal{E}_v(\mathbf{O}_{t+1})$ and $\mathbf{x}_0^a = \mathbf{A}_{t+1}$ denote the visual and action targets, and $\mathbf{z}_t = \mathcal{E}_v(\mathbf{o}_t)$ denote the reference latent. We construct noisy targets $\mathbf{x}_\tau^v=(1-\tau)\mathbf{x}_0^v+\tau\boldsymbol{\epsilon}^v$ and $\mathbf{x}_\tau^a=(1-\tau)\mathbf{x}_0^a+\tau\boldsymbol{\epsilon}^a$, with $\boldsymbol{\epsilon}^v,\boldsymbol{\epsilon}^a\sim \mathcal{N}(0,\mathbf{I})$, and define the flow-matching targets $\mathbf{v}_{\mathrm{vid}}=\boldsymbol{\epsilon}^v-\mathbf{x}_0^v$ and $\mathbf{v}_{\mathrm{act}}=\boldsymbol{\epsilon}^a-\mathbf{x}_0^a$. Conditioned on $\mathbf{z}_t,\mathbf{s}_t,\ell,\hat{\mathbf{g}}_i,\mu$, the policy $\pi^{\text{WAM}}_\phi$ jointly predicts $(\hat{\mathbf{v}}^v_\phi,\hat{\mathbf{v}}^a_\phi) = \pi^{\text{WAM}}_\phi(\mathbf{x}_\tau^v,\mathbf{x}_\tau^a,\tau\mid \mathbf{z}_t,\mathbf{s}_t,\ell,\hat{\mathbf{g}}_i,\mu)$ and is optimized by minimizing: 
\begin{equation}
    \mathcal{L}_{\mathrm{WAM}} = \mathbb{E}_{(\mathbf{o}_t,\ell,\hat{\mathbf{g}}_i),\tau,\boldsymbol{\epsilon}^v,\boldsymbol{\epsilon}^a}\left[\lVert \hat{\mathbf{v}}^v_\phi-\mathbf{v}_{\mathrm{vid}}\rVert_2^2 + \lVert \hat{\mathbf{v}}^a_\phi-\mathbf{v}_{\mathrm{act}}\rVert_2^2\right].\notag
\end{equation}

Notably, the predicted subgoal image $\hat{\mathbf{g}}_i$ can be injected into $\pi^{\text{WAM}}_\phi$ via a \emph{semantic route}, where $\hat{\mathbf{g}}_i$ is tokenized by the vision encoder and consumed by the reasoner branch, or a \emph{geometric route}, where $\hat{\mathbf{g}}_i$ is VAE-encoded into a clean latent and fed directly to the generator branch alongside the reference latent $\mathbf{z}_t$. 
We adopt the \emph{geometric route} as it grounds the subgoal at the pixel level rather than through a coarser semantic embedding (see \cref{ablation} for elaboration).  
$\hat{\mathbf{g}}_i$ is therefore encoded into $\mathbf{z}_i^g = \mathcal{E}_v(\hat{\mathbf{g}}_i)$ with the same VAE encoder $\mathcal{E}_v$ used for $\mathbf{o}_t$, while the instruction $\ell$ continues to condition the policy through the reasoner branch. We also add $\mathbf{z}_i^g$ with a zero-initialized learnable embedding to distinguish it from other latent tokens in the generator branch.

\subsectionreducemargin{In-Context Task Specification via a Global Goal} \label{method_C}

We instantiate the inference-time context $\mathcal{C}$ in \cref{icl-prelim} as a global goal image $\mathbf{G}\in \mathcal{I}$, the terminal world state that the compositional task $T$ should produce. As shown in~\cref{fig:arch}, the context enters the subgoal planner alone, which predicts $\hat{\mathbf{g}}_i \sim \mathcal{G}_\theta(\cdot \mid \mathbf{z}_t, \ell, \mathbf{G})$ at subtask $T_i$ where $\mathbf{z}_t = \mathcal{E}_v(\mathbf{o}_t)$ denotes the reference latent. 
During training, \(\mathbf{G}\) is the terminal frame of each demonstration trajectory, and the planning objective for $\mathcal{G}_\theta$ can be written as: 
% \begin{equation}
%     \mathcal{L}_{\mathrm{Plan-ICL}} = \mathbb{E}_{\tau,\boldsymbol{\epsilon}}\left[\lVert \mathcal{G}_\theta(\mathbf{x}_\tau^g,\tau \mid \mathbf{z}_t, \ell, \mathbf{G})-\mathbf{v}_{\mathrm{plan}}\rVert_2^2\right]
% \end{equation}
\begin{equation}
\mathcal{L}_{\mathrm{Plan-ICL}}
=
\mathbb{E}_{(\mathbf{o}_t,\ell,\mathbf{g}_i,\mathbf{G}),\tau,\boldsymbol{\epsilon}}
\left[
\left\|
\sqrt{\mathbf{w}_i}\odot
\left(
\mathcal{G}_{\theta}
\left(
\mathbf{x}_{\tau}^{g},\tau
\mid
\mathbf{z}_t,\ell,\mathbf{G}
\right)
-
\mathbf{v}_{\mathrm{plan}}
\right)
\right\|_2^2
\right],
\label{plan-loss-icl}\notag
\end{equation}
where \(\mathbf{w}_i\) is the end-effector ROI weight defined in~\cref{roi-weight}. At inference, \(\mathbf{G}\) is replaced by the user-specified global goal image.
The downstream policy $\pi^{\text{WAM}}_\phi(\cdot|\mathbf{z}_t,\mathbf{s}_t,\ell,\hat{\mathbf{g}}_i,\mu)$ remains unchanged as it conditions only on $\hat{\mathbf{g}}_i$, through which the effect of $\mathbf{G}$ is already mediated.
Therefore, under our framework, a global goal image unseen during training can induce a task decomposition $(T_1,\cdots,T_n)$ that was never demonstrated, with $\theta$ and $\phi$ left unchanged.

%% file: Secs/6_experiments.tex
\sectionreducemargin{Experiments} \label{Sec: experiment}
\subsectionreducemargin{Training Configuration} 

\textbf{Implementation.} The subgoal planner $\mathcal{G}_\theta$ and subgoal-guided world-action model $\pi^{\text{WAM}}_\phi$ of ViGAR are initialized from the pretrained Cosmos3-Nano checkpoint~\citep{agarwal2026cosmos}, a 16B-A8B-parameter omnimodal world model that adopts a dual-branch mixture-of-transformers architecture. 
The subgoal planner $\mathcal{G}_\theta$ is adapted from the pretrained backbone by transforming its image-to-video generation capability into image editing.
The policy $\pi^{\text{WAM}}_\phi$ is post-trained following the standard recipe for adapting world models into robot policies~\citep{ye2026world, agarwal2026cosmos}. More implementation details are listed in \cref{implement_details}.

\subsectionreducemargin{Results on Simulation Benchmark}\label{results_on_simulation}

\textbf{Simulation Benchmark.} We evaluate the compositional manipulation ability of ViGAR on the RoboTwin Clean2Random benchmark~\citep{chen2025robotwin} using the Aloha AgileX dual-arm embodiment. 
We adopt a multi-task training setup where the policy is trained exclusively on 2500 \emph{Clean} expert demonstrations~\citep{yuan2026qwen}.
All models are evaluated on 100 rollouts per task under both Clean and Random conditions.

\textbf{Baselines.} For visual subgoal prediction quality, we benchmark against VISTA~\citep{long2026scaling} and RxBrain~\citep{liang2026rxbrain}.
For policy performance, we compare ViGAR against strong baselines spanning two categories. For VLA baselines, we evaluate against StarVLA~\citep{community2026starvla}, Abot-M0~\citep{yang2026abot}, X-VLA~\citep{zheng2026x}, and $\pi_{0.5}$~\citep{intelligence2025pi_}. For WAM baselines, we evaluate against Fast-WAM~\citep{yuan2026fast}, LingBot-VA~\citep{li2026causal}, and 4D-WAM~\citep{yang20264d}. We further include a controlled baseline, Cosmos3-Nano-RoboTwin, obtained by post-training Cosmos3-Nano on the same 2500 \emph{Clean} RoboTwin trajectories with the same recipe as $\pi^{\text{WAM}}_\phi$ but without subgoal guidance, thereby isolating the contribution of subgoal guidance from that of the pretrained backbone.

\textbf{Visual Subgoal Prediction Quality.}
Before evaluating manipulation success rate, we assess the quality of the
visual subgoals generated by $\mathcal{G}_\theta$ on RoboTwin. We compare
the generated subgoals with their ground-truth targets using LPIPS, DINO cosine similarity (DINO-cos), Change of IoU ($\Delta$IoU), and retrieval correctness (RetAcc). Detailed definitions of these metrics are provided in \cref{additional_exp_subgoal_quality}.
As shown in~\cref{tab:subgoal_metrics}, ViGAR achieves the best performance across all metrics on both RoboTwin and real-robot experiments. These results show that ViGAR generates visually faithful,
semantically aligned, and subtask-correct intermediate targets.

\textbf{Policy Performance.}
As shown in \cref{tab:robotwin}, ViGAR achieves the highest success rates among the compared methods under both the \emph{Clean} (82.00\%) and \emph{Random} (67.02\%) settings. The performance boost mostly comes from the \emph{Random} setting, whereas \emph{Clean} performance is largely saturated across WAMs. The benefit of explicit subgoal selection therefore materializes precisely when the scene layout can no longer be memorized from demonstrations.
Notably, Cosmos3-Nano-RoboTwin, which follows the same policy adaptation recipe as ViGAR but omits explicit subgoal guidance, reaches only 22.46\% under \emph{Random} setting. The performance gap suggests that ViGAR benefits from predicting visual subgoals and conditioning action generation on them, validating the effectiveness of our method.

\begin{table*}[t]
    \centering
    \caption{Simulation manipulation performance and visual subgoal prediction quality of ViGAR. \textbf{Bold}: best, \underline{Underline}: second best.}
    \label{tab:main_results}
    \begin{subtable}[t]{0.494\textwidth}
        \centering
        \caption{Success rate (\%) on the RoboTwin
        Clean2Random benchmark. Methods are grouped by
        policy paradigm. }
        \label{tab:robotwin}
        \resizebox{\linewidth}{!}{%
        \begin{tabular}{llccc}
            \toprule
            Type & Model & Clean & Random & Avg. \\
            \midrule
            \multirow{4}{*}{VLA}
            & StarVLA & 46.50 & 3.20 & 24.85 \\
            & Abot-M0 & 57.40 & 30.40 & 43.90 \\
            & X-VLA & 68.00 & 20.90 & 44.45 \\
            & $\pi_{0.5}$ & 70.70 & \underline{46.00} & 58.35 \\
            \midrule
            \multirow{6}{*}{WAM}
            & Fast-WAM & 77.80 & 1.90 & 39.85 \\
            & LingBot-VA & 80.70 & 34.60 & 57.65 \\
            & 4D-WAM & \underline{81.50} & 41.80 & \underline{61.65} \\
            & \makecell[l]{Cosmos3-Nano\\-RoboTwin}
                & 77.06 & 22.46 & 49.76 \\
            & \textbf{ViGAR} & \textbf{82.00} & \textbf{67.02} & \textbf{74.51} \\
            \bottomrule
        \end{tabular}%
        }
    \end{subtable}
    \hfill
    \begin{subtable}[t]{0.492\textwidth}
        \centering
        \caption{Quality of visual subgoal prediction on the RoboTwin Clean2Random and real-world robot benchmark.}
        \label{tab:subgoal_metrics}
        \renewcommand{\arraystretch}{1.1}
        \resizebox{\linewidth}{!}{%
        \begin{tabular}{lcccc}
            \toprule
            Model & \makecell{LPIPS\\($\downarrow$)} & \makecell{DINO-cos\\($\uparrow$)} & \makecell{$\Delta\mathrm{IoU}$\\($\uparrow$)} & \makecell{RetAcc\\($\uparrow$)} \\
            \midrule
            \multicolumn{5}{l}{\textit{RoboTwin}} \\
            VISTA & \underline{0.38} & \underline{0.54} & \underline{0.22} & \underline{0.82} \\
            RxBrain & 0.73 & 0.19 & 0.18 & 0.79 \\
            \textbf{ViGAR} $\mathcal{G}_{\theta}$
                & \textbf{0.13} & \textbf{0.80} & \textbf{0.65} & \textbf{0.94} \\
            \midrule
            \multicolumn{5}{l}{\textit{Real Robot}} \\
            VISTA & \underline{0.38} & \underline{0.61} & \underline{0.24} & \underline{0.31} \\
            RxBrain & 0.60 & 0.29 & 0.18 & 0.22 \\
            \textbf{ViGAR} $\mathcal{G}_{\theta}$
                & \textbf{0.11} & \textbf{0.81} & \textbf{0.66} & \textbf{0.76} \\
            \bottomrule
        \end{tabular}%
        }
    \end{subtable}
    \vspace{-10pt}
\end{table*}

\subsectionreducemargin{Compositional Manipulation in Real-world Robot Deployment}

\textbf{Real-robot Experiment Setup.}
We conduct real-robot experiments on an AgiBot A2 robot, using its two 7-DoF arms, two wrist cameras, and one front-facing camera.
For this setting, ViGAR is first mid-trained on $\sim$900 hours of teleoperation data collected from the same embodiment, and then post-trained on task-specific real-robot demonstrations.
We evaluate ViGAR on five long-horizon compositional manipulation tasks.
Each task consists of multiple subtasks that must be completed sequentially. The tasks are as follows:
\begin{itemize}
    \item \emph{Pile Paper Cups}: pick up paper cups and stack them
    into one pile. This task contains three subtasks.
    \item \emph{Stack Colored Bowls}: organize bowls of multiple colors into separate stacks, with bowls of the same color grouped together. This task contains three subtasks.
    \item \emph{Stamp Over Letters}: stamp a letter and hand it to a person. This task contains four subtasks.
    \item \emph{Tidy-up Desktop}: collect the scattered
    trash and objects from the desktop and place them in a trash bin. This task contains four subtasks.
    \item \emph{Take-out Steam Buns}: retrieve one steamed bun from each tier
    of a two-tier bamboo steamer and place the buns on a plate. This task
    contains five subtasks.
\end{itemize}
A visualization of the task processes is provided in~\cref{fig:real_robot_task_visualize}. Each task is evaluated on 20 rollouts, and we report average task progress for each task to evaluate models' performances.

\textbf{Baselines.} We compare ViGAR with two baselines:
the VLA policy $\pi_{0.5}$ and Cosmos3-Nano-Policy.
Cosmos3-Nano-Policy is trained using the same real-robot data and policy
adaptation recipe as ViGAR, but without explicit subgoal guidance.
This comparison allows us to examine the contribution of visual subgoal planning under the same real-robot training setting.
% \vspace{-4pt}

\textbf{Visual Subgoal Prediction Quality.} We similarly assess the visual subgoals generated from real-robot observations using the same metrics and baselines as in~\cref{results_on_simulation}. As shown in~\cref{tab:subgoal_metrics}, ViGAR achieves the best performance across all metrics, consistent with its results on RoboTwin.

\begin{figure}[t]
    \centering
    \includegraphics[width=\columnwidth]{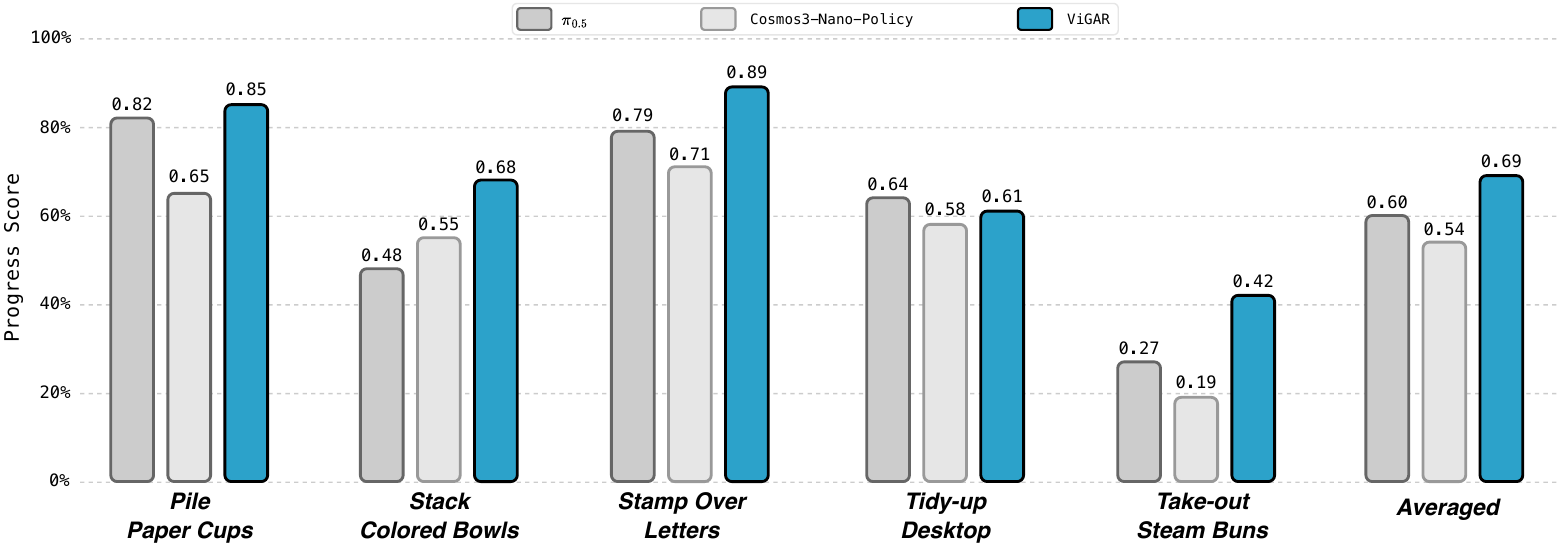}
    \caption{
    \textbf{Real-world deployment results.} We evaluate ViGAR on five long-horizon compositional manipulation tasks. Our method outperforms all baselines on the averaged task progress score.
    }
    \label{fig:real-robot-chart}

\end{figure}

\textbf{Policy Performance.}
As shown in~\cref{fig:real-robot-chart}, ViGAR achieves the best overall performance across the five real-robot tasks, outperforming both baseline policies in task progress score.
The results suggest that explicit visual subgoals provide useful task-level guidance for executing long-horizon manipulation sequences in the physical world. The comparison with Cosmos3-Nano-Policy further indicates that the improvement
is associated with subgoal-guided action generation, beyond the effect of the pretrained world-model backbone.

\subsectionreducemargin{Steering Real-world Robot Manipulation via In-Context Learning}

To examine whether a global goal image can steer ViGAR toward task
compositions observed or not observed during training, we have designed two tasks to test this capability:

\begin{itemize}
    \item \emph{Fruit Arrangement}: the robot is required to place a set of fruits into two trays, with the type and the number of fruits in each tray specified by a global goal image. 
    \item \emph{Desktop Item Storage:} the robot is required to place a set of desktop items into one storage box, with the type of items in the box specified by a global goal image.
\end{itemize}

The training demonstrations cover only a restricted set of placement patterns, while the test-time global goal image may provide an unseen pattern.
We hold fixed the initial scene and language instruction, and provide a global goal image to specify the desired task composition. This visual context is provided to the subgoal planner $\mathcal{G}_\theta$, whose predictions guide the downstream policy $\pi^{\text{WAM}}_\phi$.

\begin{figure*}[t]
    \centering
    \begin{subfigure}[b]{0.70\textwidth}
        \centering
        \includegraphics[width=\linewidth]{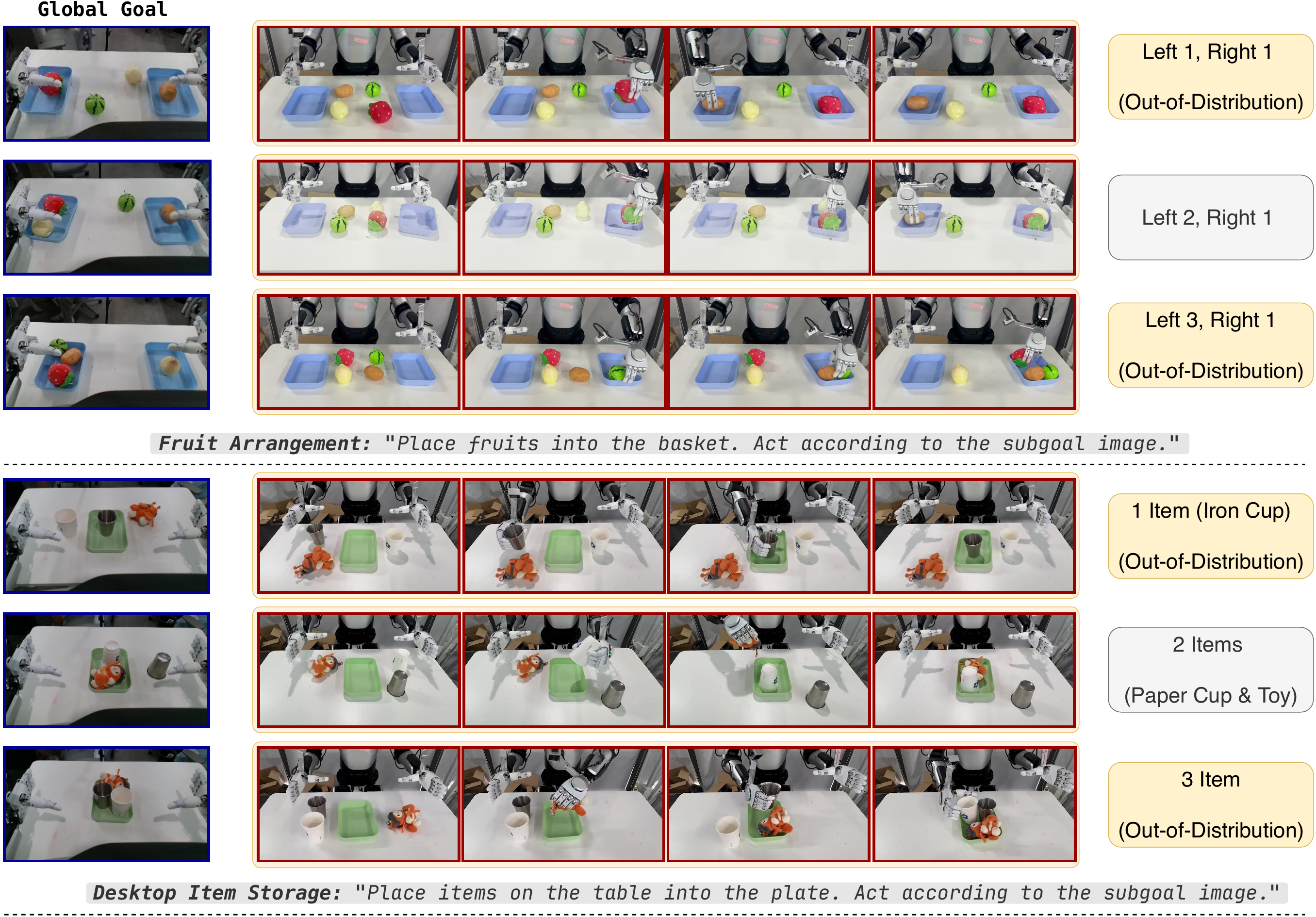}
        \caption{\textbf{Qualitative results.} Successful executions of \emph{Fruit Arrangement}
        and \emph{Desktop Item Storage} performed by ViGAR. Left column (blue border): global goal image; right columns (red border): task progress.}
        \label{fig:icl_planner_decomposition}
    \end{subfigure}
    \hfill
    \begin{subfigure}[b]{0.29\textwidth}
        \centering
        \includegraphics[width=\linewidth]{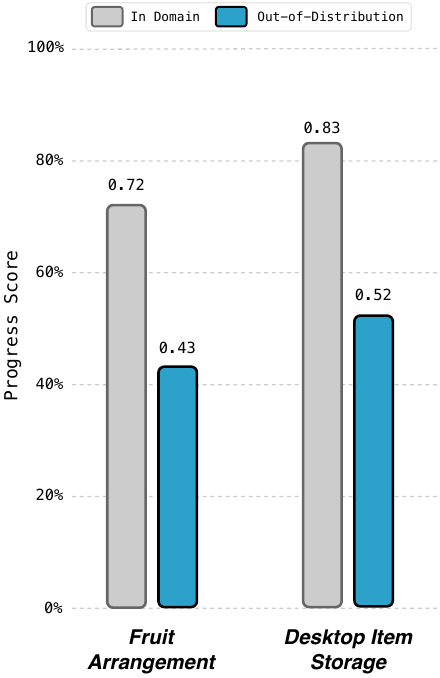}
        \caption{\textbf{Task progress scores} of ViGAR under in-domain and out-of-distribution
        global goals, each evaluated over 20 rollouts.}
        \label{fig:icl_task_nums}
    \end{subfigure}
    \caption{Results of two in-context learning tasks in real-world robot deployment: (a) qualitative execution
    sequences and (b) task progress scores on in-domain and out-of-distribution settings.}
    \label{fig:icl_results}
    % \vspace{-10pt}
\end{figure*}

As shown in~\cref{fig:icl_planner_decomposition,fig:icl_task_nums}, ViGAR executes the placements specified by the global goal, achieving average task progress score of 77.5\% (in-domain) and 47.5\% (OOD). This shows that visual context can guide subtask selection and composition, validating that
learned manipulation skills can be composed into a new task through an inference-time specification of the desired terminal state.

% \vspace{-4pt}
\subsectionreducemargin{Ablation Studies} \label{ablation}
\textbf{Effectiveness of Implementation Design.} We investigate the effectiveness of leveraging subgoal look-ahead rule (\cref{look-ahead rule}) in training $\mathcal{G}_\theta$ and $\pi_{\phi}^{\mathrm{WAM}}$, and end-effector ROI-weighted flow matching loss (\cref{plan-loss}) in training $\mathcal{G}_\theta$. 
As shown in~\cref{tab:ablation_subgoal_lookahead_ROI_reweighted}, removing either component degrades performance under both \emph{Clean} and \emph{Random} settings, validating the effectiveness of our implementation design.
Specifically, end-effector ROI weighting contributes most to the performance gain, as it concentrates supervision more on manipulation-related regions than on the static background.
Subgoal look-ahead also brings improvement by preventing regressing back to subtasks that have already been completed.

\textbf{Goal Injection Route.} To isolate the effect of the injection route, we use ground-truth goal images in this controlled ablation rather than predicted ones. Alongside a baseline without goal guidance, we compare three injection schemes: vision-encoder tokens provided to the reasoner (\emph{semantic route}), VAE latents provided to the generator (\emph{geometric route}), and both routes combined. For a fair comparison, all variants are trained for 30k steps on the same 2500 RoboTwin \emph{Clean} trajectories.

As shown in~\cref{tab:goal_injection}, generator-only injection achieves the highest average success rate. Adding the reasoner route improves performance on \emph{Clean} but reduces robustness under \emph{Random} conditions. These results motivate the geometric route as the goal
injection mechanism used in ViGAR.

\begin{table*}[t]
    \centering
    \caption{Ablation studies on the implementation of ViGAR. Success rates (\%) over 50 RoboTwin tasks; \textbf{Bold}: best.}
    \label{tab:goal_ablations}
    \small
    \renewcommand{\arraystretch}{1.1}

    \begin{subtable}[t]{0.47\textwidth}
        \centering
        \caption{Ablation on the choice of goal injection routes using ground-truth goal images.}
        \label{tab:goal_injection}
        \resizebox{\linewidth}{!}{%
        \begin{tabular}{lccc}
            \toprule
            Goal injection & Clean & Random & Avg. \\
            \midrule
            None                 & 76.58 & 21.78 & 49.18 \\
            Reasoner             & 78.42 & 23.52 & 50.97 \\
            Generator            & 78.92 & \textbf{45.42} & \textbf{62.17} \\
            Generator + Reasoner & \textbf{82.16} & 41.28 & 61.72 \\
            \bottomrule
        \end{tabular}}
    \end{subtable}
    \hfill
    \begin{subtable}[t]{0.50\textwidth}
        \centering
        \caption{Ablation on the effectiveness of subgoal look-ahead rule and end-effector ROI-weighted training.}
        \label{tab:ablation_subgoal_lookahead_ROI_reweighted}
        \resizebox{\linewidth}{!}{%
        \begin{tabular}{lccc}
            \toprule
            Subgoal supervision  & Clean & Random & Avg. \\
            \midrule
            ViGAR                & \textbf{82.00} & \textbf{67.02} & \textbf{74.51} \\
            \quad w/o subgoal look-ahead & 79.80 & 65.92 & 72.86 \\
            \quad w/o EEF ROI weighting & 79.28 & 65.64 & 72.46 \\
            \quad w/o both       & 72.68 & 52.40 & 62.54 \\
            \bottomrule
        \end{tabular}}
    \end{subtable}
    % \vspace{-15pt}
\end{table*}

%% file: Secs/7_conclusion.tex
\sectionreducemargin{Conclusion}
\label{conclusion}

In this work, we addressed the gap between short-horizon world-action prediction and task-level planning for compositional manipulation.
We introduced ViGAR, which factorizes manipulation into visual subgoal planning and subgoal-conditioned joint video-action generation.
By representing intermediate targets as images, ViGAR connects task-level decisions with the physical dynamics modeled by world-action models.
Experiments on RoboTwin and five real-robot tasks support the effectiveness of this design. Furthermore, a global goal image provides an interface for inference-time steering, enabling ViGAR to compose learned manipulation behaviors into tasks absent from the training demonstrations without parameter updates. 
These findings suggest that explicit visual goals offer a useful connection between planning and action generation, providing a promising direction for compositional and steerable robotic manipulation.

% \section*{AI Use Statement}
% In this work, generative AI tools were used to assist with polishing of the manuscript, synthetic dataset generation, and method implementation.
% We take responsibility for the final content of this work, including text, claims or artifacts produced with the aid of generative AI.

%% file: Secs/8_appendix.tex
\appendix

\startcontents[chapters]

\begin{center}
  \textbf{\Large Rethinking World-Action Model for Compositional \\ and In-Context Robotic Manipulation} \vspace{0.5cm} \\
  {\Large Appendix}
  \vspace{0.5cm}
\end{center}

\printcontents[chapters]{}{1}{}

\section{Dataset}\label{dataset}

\textbf{Simulation Data.} 
For simulation experiments, each demonstration is divided into annotated stages, where an input observation is paired with the expert image at the end of its target stage (i.e., subgoal image), the corresponding instruction, proprioceptive state and action trajectories. 
We construct subtask annotations for the 2,500 clean expert rollouts provided by RoboTwin~\citep{chen2025robotwin} according to the compositional structure of each task. The following 19 tasks are treated as long-horizon and segmented into multiple subtasks: 

\emph{beat block hammer}, \emph{blocks ranking rgb},  \emph{blocks ranking size}, \emph{dump bin bigbin}, \emph{handover mic}, \emph{hanging mug}, \emph{move playingcard away}, \emph{open laptop}, \emph{place bread basket}, \emph{place can basket}, \emph{place cans plasticbox}, \emph{place dual shoes}, \emph{place phone stand}, \emph{put bottles dustbin}, \emph{put object cabinet}, \emph{stack blocks three}, \emph{stack blocks two}, \emph{stack bowls three}, \emph{stamp seal}.

For these tasks, the endpoint frame of each subtask is used as its ground-truth subgoal image. The remaining 31 tasks are identified as single-stage tasks with one subgoal image, namely the terminal frame of the trajectory. This preprocessing provides subtask-level supervision without changing the original expert rollouts.

\textbf{Real-Robot Data.} 
Our real-robot dataset comprises approximately 900 hours of teleoperated demonstrations collected using the AgiBot A2 platform. It covers 752 tasks involving more than 200 distinct objects across over 10 diverse backgrounds. All episodes are annotated through a human-in-the-loop pipeline, in which each trajectory is manually segmented into subtasks corresponding to atomic robotic skills and aligned with the semantic structure of the overall task.

\section{Implementation Details}\label{implement_details}

\textbf{Implementation of Subgoal Planner $\mathcal{G}_\theta$.} Cosmos3 has no native support for image editing, so we adapt its image-to-video capability to image editing as follows: the current observation $\mathbf{o}_t$ is encoded as a single clean conditioning latent frame $\mathbf{z}_t$, while the target subgoal image $\mathbf{g}_i$ is replicated four times before VAE encoding, yielding a single target latent frame. Multi-view observations and subgoal images are concatenated along the frame dimension prior to VAE encoding. During training of $\mathcal{G}_\theta$, we freeze the reasoner branch and update only the generator-side parameters.

We use two planner variants, termed \emph{Base} and \emph{ROI}, both initialized from Cosmos3-Nano and trained for 100k steps on a large corpus of robot manipulation data spanning single-arm and bimanual embodiments~\citep{o2024open,khazatsky2024droid,bu2025agibot,wu2025robocoin,tian2026interndata}.
\emph{Base} uses the standard flow-matching objective for image-editing training throughout, whereas \emph{ROI} uses the standard objective for the first 30k steps, followed by 70k steps with end-effector region-of-interest weighting. The latter emphasizes the regions around the projected end-effectors in the target image, using a foreground-to-background weight ratio of $4{:}1$. Both variants use AdamW with a learning rate of $2\times10^{-5}$ (50-step warm-up, then constant) and a global batch size of 32. The generator branch and its visual projections are trainable, while the reasoner branch and visual tokenizer remain frozen.

% \vspace{-4pt}
\textbf{Implementation of World-Action Model $\pi_\phi^{\mathrm{WAM}}$.} $\pi_\phi^{\text{WAM}}$ is adapted from the same Cosmos3-Nano backbone following the Cosmos3 robot-policy post-training recipe~\citep{ye2026world, agarwal2026cosmos}: an action encoder $\mathcal{E}_a$, an action decoder $\mathcal{D}_a$, and an action-modality embedding are introduced and zero-initialized for the target embodiment, along with the zero-initialized goal-role embedding used to tag subgoal latents (\cref{method_B}). Multi-view observations and subgoal images follow the same concatenation scheme as in $\mathcal{G}_\theta$. 
We set the action chunk horizon $H_a=48$ and video horizon $H_v=48$ with a video downsample rate of 4, so that $\pi_\phi^{\mathrm{WAM}}$ jointly predicts a chunk of 48 actions and 12 video frames for each control query.
To alleviate the train-test mismatch, we apply the trained $\mathcal{G}_\theta$ to the
demonstrations used to train $\pi^{\text{WAM}}_\phi$ and obtain predicted subgoals
$\hat{\mathbf{g}}_i$. Predictions that are semantically inconsistent or visually distorted are
replaced by the corresponding ground-truth subgoals $\mathbf{g}_i$, and $\pi^{\text{WAM}}_\phi$
is trained on the resulting mixture of predicted and ground-truth subgoals. 

During training, the reasoner branch remains frozen, while the generator branch, modality embedding, action projections, and goal-role embedding are trainable. Our policy is trained for 50k steps with a global batch size of 256. We use a base learning rate of $1\times 10^{-5}$, with $10\times$ learning rates for the action projections and action embedding and $5\times$ for the goal embedding. 
All learning rates follow a linear schedule over a 100k-step cycle without warm-up.
We also apply ColorJitter during policy training with brightness, contrast, and saturation strengths of $B=0.3$, $C=0.4$, and $S=0.5$, respectively. Augmentation is disabled at inference.

\textbf{Subtask transition of ViGAR.} At deployment, ViGAR runs in a receding-horizon loop: at each control query, \(G_\theta\) recomputes \(\hat{\mathbf{g}}_i\) from the current observation (and optional \(\mathbf{G}\)), then \(\pi_\phi^{\mathrm{WAM}}\) predicts one action chunk. This implicitly advances the subtask as the scene evolves, with no external detector of subtask completion.

\begin{table}[htbp]
\centering
\caption{Training configurations and hyperparameters of ViGAR.}
\label{tab:robotwin-settings}
\small
\setlength{\tabcolsep}{5pt}
\scalebox{0.9}{%
\begin{tabular}{@{}>{\raggedright\arraybackslash}p{0.28\linewidth}>{\raggedright\arraybackslash}p{0.33\linewidth}>{\raggedright\arraybackslash}p{0.33\linewidth}@{}}
\toprule
\textbf{Setting} & \textbf{Subgoal planner} & \textbf{Policy} \\
\midrule
Pretrained backbone & Cosmos3-Nano & Cosmos3-Nano \\
Camera views & Head + two wrists & Head + two wrists \\
Image size & $384\times320$ & $352\times288$ \\
Optimizer & AdamW & FusedAdam \\
Adam $(\beta_1,\beta_2)$ & $(0.9,0.95)$ & $(0.9,0.99)$ \\
Base LR & $2\times10^{-5}$ & $1\times 10^{-5}$ \\
LR schedule & 50-step warm-up, then constant & Linear decay \\
Global batch size & 32 & 256 \\
Weight decay & 0 & 0.05 \\
Gradient clipping & 0.1 & 1.0 \\
Precision & BF16 & BF16 \\
% ColorJitter $(B,C,S,H)$ & --- & $(0.3,0.4,0.5,0)$ \\
\bottomrule
\end{tabular}}
\end{table}
% \vspace{-6pt}

\textbf{Inference Infrastructure Optimization.} For the policy $\pi_\phi^{\text{WAM}}$, we apply a cascade of numerically lossless optimizations: 
(i)~static compilation with CUDA graph capture over the Transformer region eliminates kernel-launch overhead;
(ii)~fused QKV and SwiGLU projections reduce per-chunk GEMM dispatches;
(iii)~batched CFG packs the conditional and unconditional inputs into a single
forward pass with independent attention boundaries, halving the number of network evaluations;
(iv)~exact condition caching reuses the goal-frame latent and text tokens by content hash when the subgoal has not changed. 
For the subgoal planner $\mathcal{G}_\theta$, the same optimization stack adopted for $\pi_\phi^{\text{WAM}}$ is applied. We additionally adopt TeaCache~\citep{liu2025timestepembeddingtellsits}, which skips Transformer evaluations at diffusion steps whose residual change falls below a relative-$\ell_1$ threshold. 
These optimizations accelerate policy inference and subgoal image generation by $2.5\times$ and $11.2\times$ on a single NVIDIA B20Z GPU, respectively.
The inference optimization substantially improves the usability and instantaneity of ViGAR especially in real-world deployment.

\section{Additional Experiment Results}\label{additional_exp}
\subsectionreducemargin{Visualization of Real-world Robot Task Settings}

\begin{figure}[htbp]
    \centering
    \includegraphics[width=0.82\columnwidth]{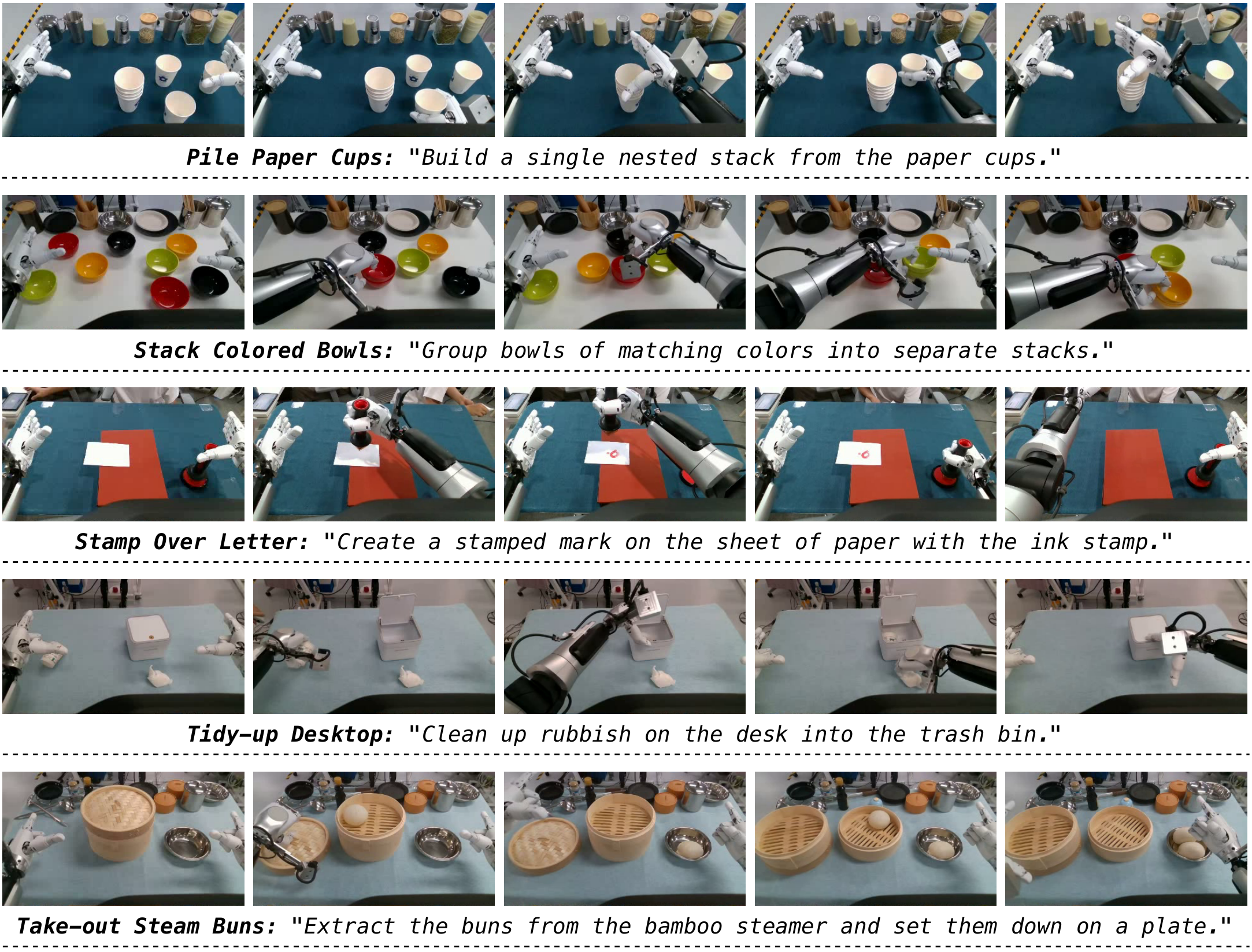}
    \caption{
    \textbf{Visualization of the five real-world compositional manipulation tasks.} The figure displays the task progress of \emph{Pile Paper Cups, Stack Colored Bowls, Stamp over Letters, Tidy-up Desktop}, and \emph{Take-out Steam Buns}.
    }
    \label{fig:real_robot_task_visualize}

    \centering
    \includegraphics[width=0.82\columnwidth]{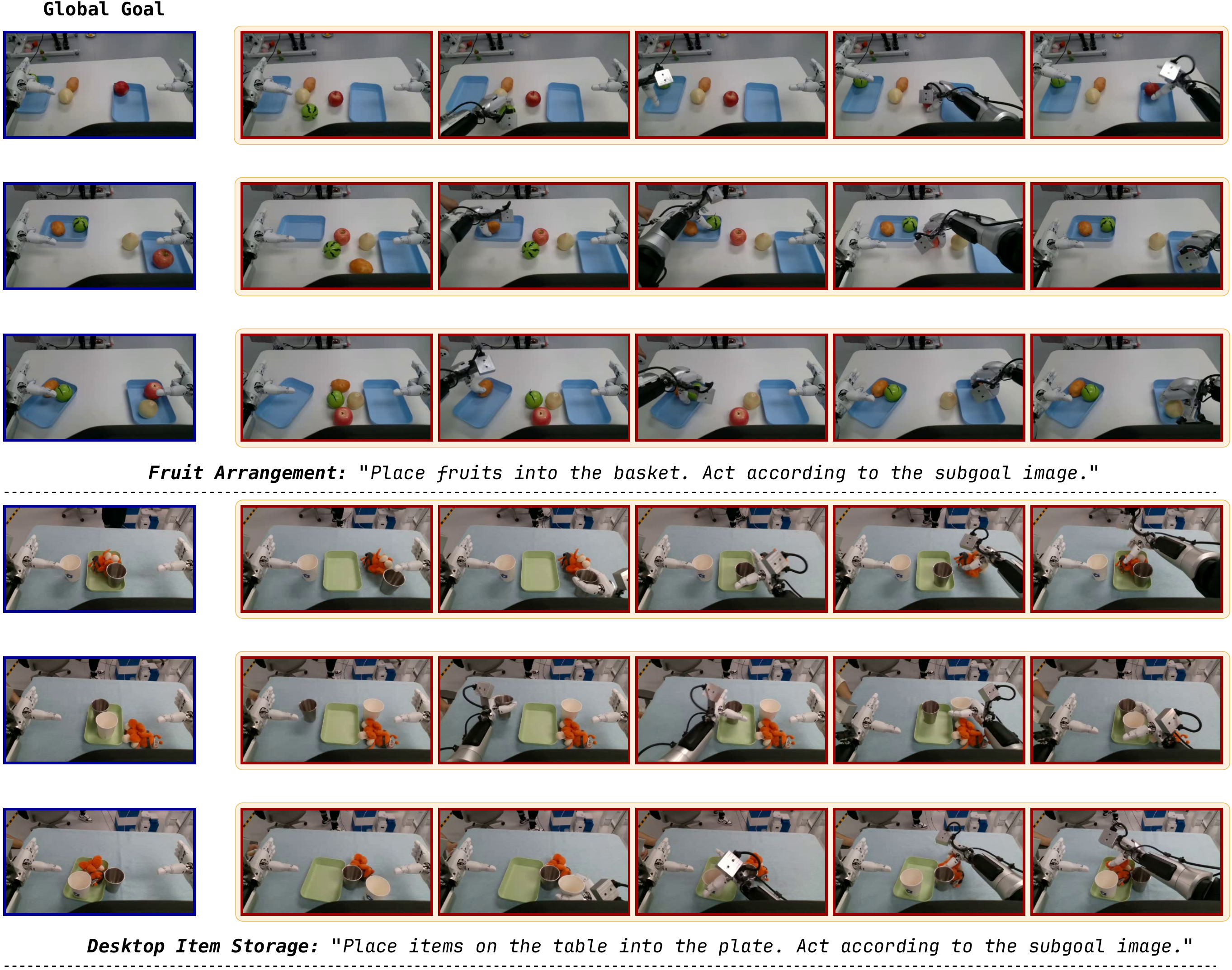}
    \caption{
    \textbf{Visualization of the two real-world in-context learning tasks.} The figure displays the task progress of \emph{Fruit Arrangement} and \emph{Desktop Item Storage}.
    Left Column (blue border): global goal image of the task; Right Columns (red border): task progress of in-context learning tasks.
    }
    \label{fig:icl_task_visualize}
\end{figure}

\newpage

\subsection{Visual Subgoal Prediction Quality}
\label{additional_exp_subgoal_quality}

We evaluate the subgoal planner independently of the downstream action policy.
For each reference observation $o_t$, the planner predicts a subgoal image
$\hat{g}_i$, which is compared with the corresponding ground-truth subgoal
$g_i$. All methods use the same reference observations and evaluation targets
on both RoboTwin and the real-robot benchmark.

We use the following metrics:

\begin{itemize}
    \item \textbf{LPIPS} ($\downarrow$): perceptual distance between the
    predicted subgoal $\hat{g}_i$ and the ground-truth subgoal $g_i$,
    computed using the AlexNet-based LPIPS metric. Lower values indicate
    better perceptual similarity.
    \item \textbf{DINO cosine similarity} (DINO-cos, $\uparrow$):
    cosine similarity between the DINOv2-base CLS features of $\hat{g}_i$
    and $g_i$, measuring their semantic alignment. Higher values indicate
    stronger semantic similarity.
    \item \textbf{Change-IoU} ($\Delta\mathrm{IoU}$, $\uparrow$): IoU between the change masks of the predicted and ground-truth subgoals. The change masks are obtained by computing the grayscale difference between each target image and the input observation, followed by blurring and thresholding the difference at 25. Higher values indicate better alignment of the predicted and ground-truth spatial changes.
    \item \textbf{Retrieval accuracy} (RetAcc, $\uparrow$):
    DINOv2 features are used to retrieve the most similar image from the
    endpoint images of all subtasks in the evaluation set. RetAcc measures
    the fraction of predictions for which the retrieved endpoint corresponds
    to the ground-truth target subtask. Higher values indicate more accurate
    subtask-level prediction.
\end{itemize}

These metrics evaluate complementary aspects of visual subgoal prediction,
including perceptual fidelity, semantic alignment, spatial-change consistency,
and subtask-level correctness. The results are reported in
\cref{tab:subgoal_metrics}.

\subsection{Detailed Results on Real-robot Experiments} \label{additional_exp_real_robot}

We quantify the performance of real-robot tasks by two metrics: task progress score and stage-wise success rate.
Each long-horizon task is decomposed into $K$ sequential stages, where a stage can only be attempted after all of its preceding stages have been completed.
For each task and each method, we conduct $N=20$ independent trials.
Each trial $i$ is associated with a goal requiring $K_i$ sequential stages. For the five long-horizon tasks and the in-domain in-context settings, $K_i\equiv K$; for the out-of-distribution (OOD) in-context settings, the $N=20$ trials are split evenly between two unseen goal configurations with different $K_i$.
We denote by $s_i\in\{0,1,\dots,K_i\}$ the number of stages completed before the first failure.

\textbf{Stage-wise success rate.}
\begin{equation}
\mathrm{SR}_k=\frac{1}{N_k}\sum_{i:\,K_i\ge k}\mathbf{1}[s_i\ge k],\qquad
N_k=\bigl|\{i: K_i\ge k\}\bigr|,
\end{equation}
i.e., stage $k$ is evaluated only on trials whose goal contains at least $k$ stages.

\textbf{Task progress score.}
\begin{equation}
\mathrm{TPS}=\frac{1}{N}\sum_{i=1}^{N}\frac{s_i}{K_i}.
\end{equation}
When all trials share the same $K$ (\cref{tab:real_robot_full} and each column of~\cref{tab:real_robot_icl}), $N_k=N$ and this reduces to $\mathrm{TPS}=\frac{1}{K}\sum_{k=1}^{K}\mathrm{SR}_k$, with $\mathrm{SR}_K$ being the full-task success rate; the OOD avg.\ entries in~\cref{tab:real_robot_icl} apply the general form over all 20 OOD trials. The stage-wise success rate and task progress scores of all five long-horizon manipulation tasks and two in-context learning tasks are reported in~\cref{tab:real_robot_full} and~\cref{tab:real_robot_icl}.

\begin{table}[t]
    \centering
    \caption{Stage-wise success rates and task progress scores of ViGAR on real-robot long-horizon manipulation tasks. Each task is evaluated over 20 rollouts. \textbf{Bold}: best task progress.}
    \label{tab:real_robot_full}
    \label{tab:real_robot_stage}
    \small
    \setlength{\tabcolsep}{2pt}
    \renewcommand{\arraystretch}{1.12}
    \begin{tabular}{ @{} l c @{\hspace{7pt}} c @{\hspace{9pt}} c @{\hspace{9pt}} c @{} }
        \toprule
        \textbf{Task}
        & \textbf{Stage}
        & $\boldsymbol{\pi}_{0.5}$
        & \makecell{\textbf{Cosmos3-Nano-Policy}}
        & \textbf{ViGAR} \\
        \midrule

        \multirow[c]{4}{*}{\mbox{Pile Paper Cups}}
        & S1 & 0.95 & 0.85 & 0.90 \\
        & S2 & 0.85 & 0.65 & 0.85 \\
        & S3 & 0.65 & 0.45 & 0.80 \\
        \cmidrule(l){2-5}
        & Progress & 0.82 & 0.65 & \textbf{0.85} \\
        \midrule

        \multirow[c]{4}{*}{\mbox{Stack Colored Bowls}}
        & S1 & 0.70 & 0.75 & 0.85 \\
        & S2 & 0.50 & 0.55 & 0.70 \\
        & S3 & 0.25 & 0.35 & 0.50 \\
        \cmidrule(l){2-5}
        & Progress & 0.48 & 0.55 & \textbf{0.68} \\
        \midrule

        \multirow[c]{5}{*}{\mbox{Stamp over Letter}}
        & S1 & 1.00 & 0.95 & 1.00 \\
        & S2 & 0.80 & 0.85 & 0.95 \\
        & S3 & 0.70 & 0.60 & 0.80 \\
        & S4 & 0.65 & 0.45 & 0.80 \\
        \cmidrule(l){2-5}
        & Progress & 0.79 & 0.71 & \textbf{0.89} \\
        \midrule

        \multirow[c]{5}{*}{\mbox{Tidy-up Desktop}}
        & S1 & 0.85 & 0.80 & 0.85 \\
        & S2 & 0.70 & 0.65 & 0.65 \\
        & S3 & 0.55 & 0.50 & 0.55 \\
        & S4 & 0.45 & 0.35 & 0.40 \\
        \cmidrule(l){2-5}
        & Progress & \textbf{0.64} & 0.58 & 0.61 \\
        \midrule

        \multirow[c]{6}{*}{\mbox{Take-out Steam Buns}}
        & S1 & 0.60 & 0.50 & 0.65 \\
        & S2 & 0.35 & 0.25 & 0.55 \\
        & S3 & 0.25 & 0.15 & 0.45 \\
        & S4 & 0.15 & 0.05 & 0.30 \\
        & S5 & 0.00 & 0.00 & 0.15 \\
        \cmidrule(l){2-5}
        & Progress & 0.27 & 0.19 & \textbf{0.42} \\
        \midrule

        \multicolumn{2}{@{}l}{Average Progress} & 0.60 & 0.54 & \textbf{0.69} \\
        \bottomrule
    \end{tabular}
\end{table}

\begin{table}[t]
    \centering
    \caption{Stage-wise success rates and task progress scores of ViGAR on in-context real-robot tasks. Results are computed over 20 rollouts.}
    \label{tab:real_robot_icl}
    \small
    \setlength{\tabcolsep}{6pt}
    \renewcommand{\arraystretch}{1.12}
    \begin{tabular}{@{}llccc@{}}
        \toprule
        \multirow{2}{*}{\textbf{Task}} & \multirow{2}{*}{\textbf{Stage}} & \multirow{2}{*}{\textbf{In-domain}} & \multicolumn{2}{c}{\textbf{OOD}} \\
        \cmidrule(l){4-5}
        & & & \textbf{Setting A} & \textbf{Setting B} \\
        \midrule
        \multirow[c]{6}{*}{\mbox{Fruit Arrangement}}
            & S1       & 0.95 & 0.70 & 0.60 \\
            & S2       & 0.70 & 0.30 & 0.40 \\
            & S3       & 0.50 & --   & 0.30 \\
            & S4       & --   & --   & 0.10 \\
            \cmidrule(l){2-5}
            & Progress & 0.72 & 0.50 & 0.35 \\
            & Progress avg. & 0.72 & \multicolumn{2}{c}{0.43} \\
        \midrule
        \multirow[c]{5}{*}{\mbox{Desktop Item Storage}}
            & S1       & 0.90 & 0.70 & 0.60 \\
            & S2       & 0.75 & --   & 0.30 \\
            & S3       & --   & --   & 0.10 \\
            \cmidrule(l){2-5}
            & Progress & 0.83 & 0.70 & 0.33 \\
            & Progress avg. & 0.83 & \multicolumn{2}{c}{0.52} \\
        \bottomrule
    \end{tabular}
\end{table}

\begin{figure}[hbtp]
    \centering
    \includegraphics[width=\columnwidth]{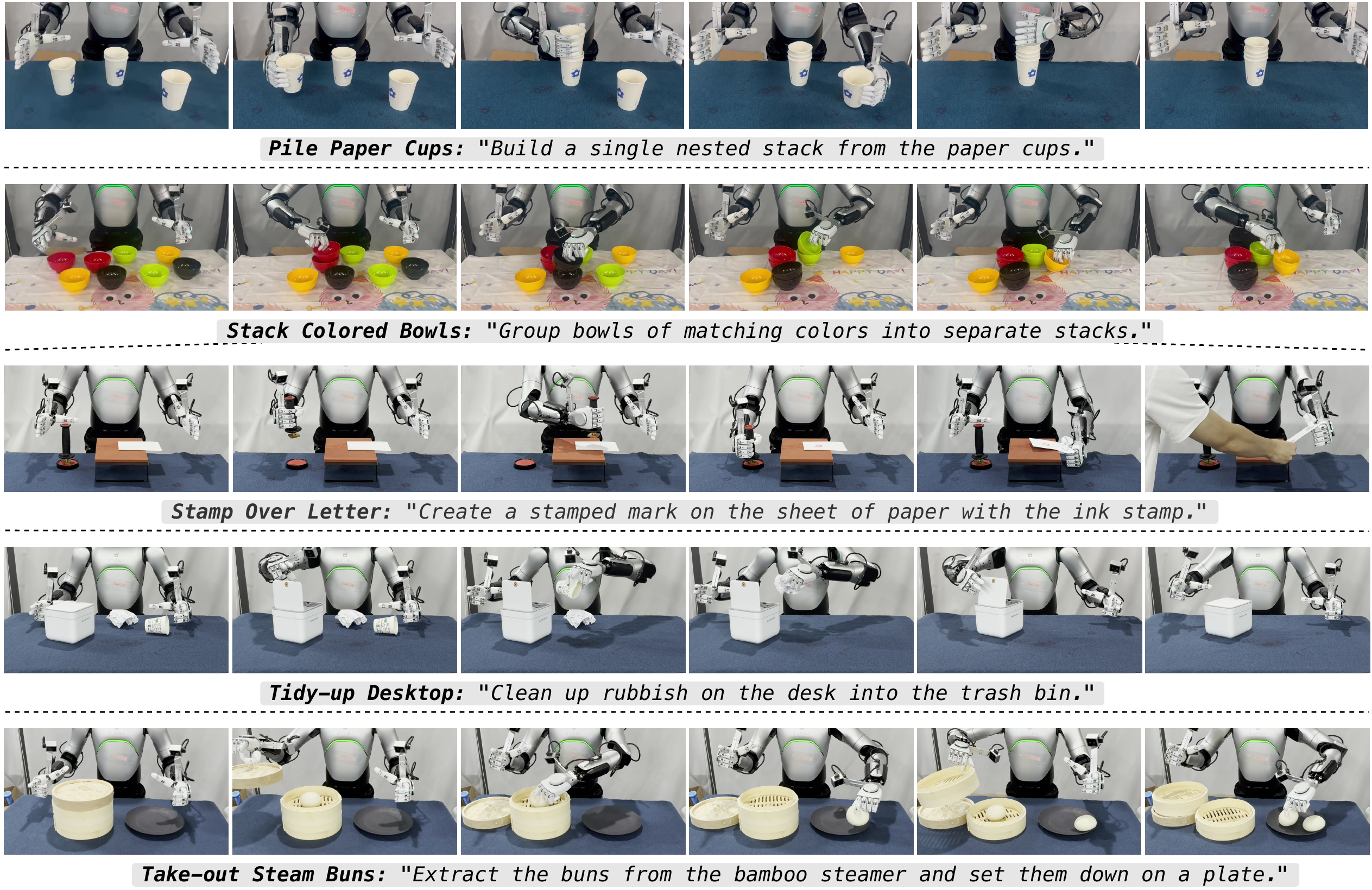}
    \caption{
    \textbf{Qualitative results of five real-world compositional manipulation tasks.} The figure displays the task progress of \emph{Pile Paper Cups, Stack Colored Bowls, Stamp Over Letter, Tidy-up Desktop}, and \emph{Take-out Steam Buns}.
    }
    \label{fig:real_robot_qualitative}
\end{figure}

We also provide the qualitative results of five real-world compositional manipulation tasks besides the task progress score illustrated in~\cref{fig:real-robot-chart}. The visualization is shown in~\cref{fig:real_robot_qualitative}.

\begin{figure}[hbtp]
    \centering
    \includegraphics[width=\columnwidth]{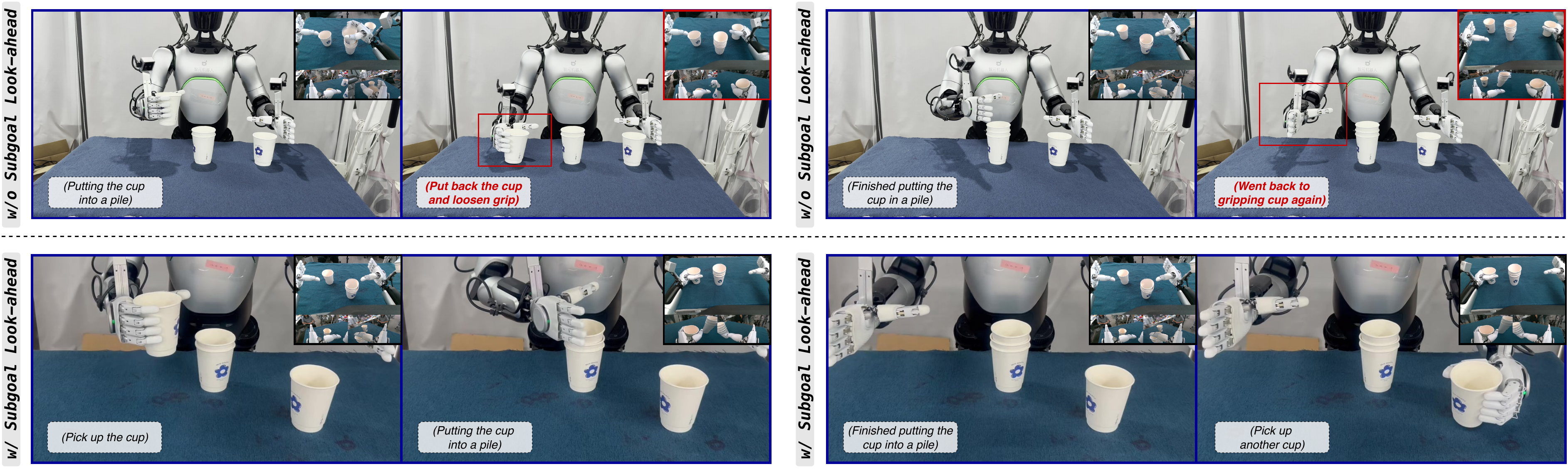}
    \caption{
    \textbf{Empirical evidence for the subgoal look-ahead rule.} Red borders indicate problematic subgoal images and the resulting manipulation behavior. ViGAR trained without subgoal look-ahead tends to regress to subtasks that have already been completed, whereas training with subgoal look-ahead alleviates this issue.
    }
    \label{fig:real_robot_subgoal_lookahead}
\end{figure}